\documentclass[lettersize,journal]{IEEEtran}
\usepackage[colorlinks=true,linkcolor=black,urlcolor=black,citecolor=black]{hyperref}
\usepackage{amsmath,amsfonts}
\usepackage{algorithmic}
\usepackage{algorithm}
\usepackage{array}
\usepackage[caption=false,font=normalsize,labelfont=sf,textfont=sf]{subfig}
\usepackage{textcomp}
\usepackage{stfloats}
\usepackage{url}
\usepackage{verbatim}
\usepackage{graphicx}
\usepackage{cite}
\usepackage{soul, color, xcolor}

\begin{document}

\title{Robotic Grinding Skills Learning Based on \\Geodesic Length Dynamic Motion Primitives}

\author{Shuai Ke, Huan Zhao, ~\IEEEmembership{Member,~IEEE}, Xiangfei Li, ~\IEEEmembership{Member,~IEEE},  Zhiao Wei,
\\ Yecan Yin, Han Ding, ~\IEEEmembership{Senior Member,~IEEE}
\thanks{Manuscript received April 24, 2024. This work was supported by the National Natural Science Foundation of China under Grant Nos. 52188102, 52090054 and 52205521, and by the Postdoctoral Fellowship Program of CPSF under Grant Number GZC20240539.}
\thanks{The authors are with the State Key Laboratory of Intelligent Manufacturing Equipment and Technology, Department of Mechanical Science and Engineering, Huazhong University of Science and Technology, Wuhan 430074, China (e-mail: keshuai@hust.edu.cn; huanzhao@hust.edu.cn; lixiangfei@hust.edu.cn; zhiao\_wei@hust.edu.cn; yinyecan6666@hust.edu.cn; dinghan@hust.edu.cn)}}

\markboth{Journal of \LaTeX\ Class Files}%
{Shell \MakeLowercase{\textit{et al.}}: A Sample Article Using IEEEtran.cls for IEEE Journals}


\maketitle

\begin{abstract}
Learning grinding skills from human craftsmen by imitation learning has emerged as a prominent research topic in the field of robotic machining. Given their robust trajectory generalization ability and resilience to various external disturbances and environmental changes, Dynamical Movement Primitives (DMPs) provide a promising skills learning solution for the robotic grinding. However, challenges arise when directly applying DMPs to grinding tasks, including low orientation accuracy, inaccurate synchronization of position, orientation, and force, and the inability to generalize surface trajectories. To address these issues, this paper proposes a robotic grinding skills learning method based on geodesic length DMPs (Geo-DMPs). First, a normalized two-dimensional weighted Gaussian kernel function and intrinsic mean clustering algorithm are proposed to extract surface geometric features from multiple demonstration trajectories. Then, an orientation manifold distance metric is introduced to exclude the time factor from the classical orientation DMPs, thereby constructing Geo-DMPs for the orientation learning to improve the orientation trajectory generation accuracy. On this basis, a synchronization encoding framework for position, orientation, and force skills is established, using a phase function related to geodesic length. This framework enables the generation of robotic grinding actions between any two points on the surface. Finally, experiments on robotic chamfer grinding and free-form surface grinding demonstrate that the proposed method exhibits high geometric accuracy and good generalization capabilities in encoding and generating grinding skills. This method holds significant implications for learning and promoting robotic grinding skills. To the best of our knowledge, this may be the first attempt to use DMPs to generate grinding skills for position, orientation, and force on model-free surfaces, thereby presenting a novel approach to robotic grinding skills learning.
\end{abstract}


\section{Introduction}
\IEEEPARstart{C}{ompared} to traditional machine tools, robots demonstrate greater flexibility in the grinding of complex parts \cite{liu2022}. However, current grinding tasks typically rely on workpiece design models combined with accumulated process knowledge from long-term experiments to plan the grinding paths and apply the appropriate grinding forces. The challenges of grinding small batches and parts with complex structures lead to substantial costs in manual programming and debugging. Therefore, incorporating human experience into grinding tasks has emerged as a research-worthy problem \cite{field2016,deng2018,yin2017}. Imitation learning combined with artificial intelligence algorithms for task modeling and learning \cite{zhu2018,yangc2018,zhang2025} has enabled the efficient programming of robots. This approach is applicable to multi-variety small-batch task scenarios, resulting in effective reductions in labor costs and setup time \cite{mukherjee2022}. The typical grinding tasks are shown in  Fig. 1.

\begin{figure}[t]
    \centering
    \includegraphics[scale=0.24, angle=0, origin=c]{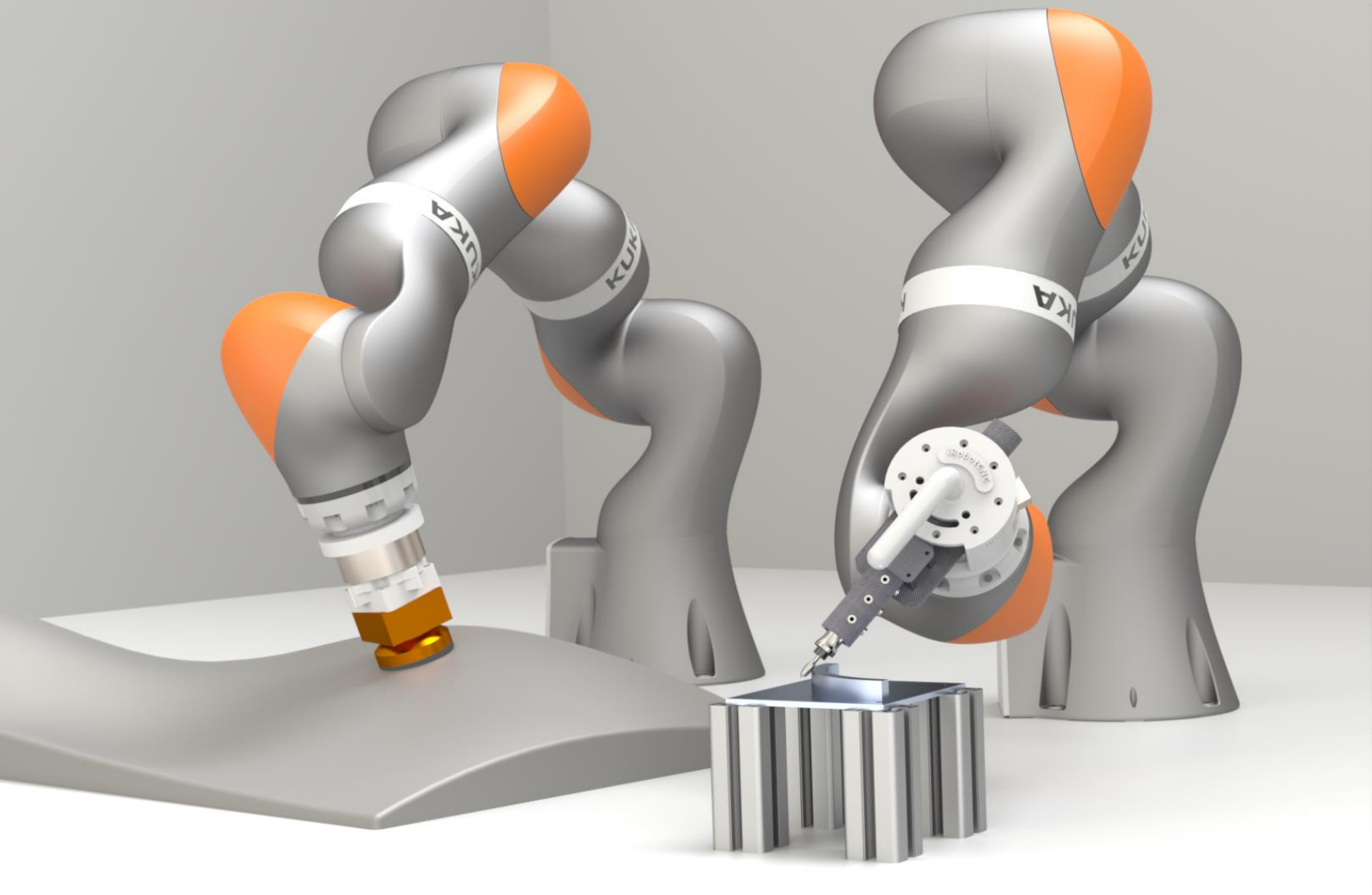}
    \caption{Example of robotic grinding skills: free-form surface and chamfer grinding.}
    \label{fig1}
\end{figure}

In recent years, imitation learning has garnered extensive research attention in robot operation programming within machining scenarios \cite{chen2022}. By capturing continuous trajectories and force characteristics of human demonstrators through visual motion capture \cite{ng2015} or directly dragging the robot \cite{franzese2023}, the basic motion primitives (MPs)\cite{giszter1993} of humans are encoded. The obtained motion primitives can be applied to similar work scenarios \cite{billard2019}. Various motion encoding methods have been proposed to adapt to different skill characteristics. Huang et al. \cite{huang2019} proposed kernelized motion primitives (KMP), which minimizes the kullback-leibler divergence between parameterized trajectories and sample trajectories to obtain a non-parametric skill learning model. There are also other methods for encoding human motion skills, such as Hidden Markov Models (HMM) \cite{rabiner1989}, Gaussian Processes (GP) \cite{williams2006}, and Probabilistic Motion Primitives (ProMP) \cite{paraschos2013}. These methods have found wide applications in obstacle avoidance \cite{lu2021}, grasping \cite{abdo2013}, and assembly \cite{su2021,ti2022}. However, when using learned trajectories for similar scenarios, the above methods exhibit limited generalization capabilities. Ijspeert et al. \cite{ijspeert2013} proposed Dynamic Motion Primitives (DMPs), which utilize a spring-damping model and force-coupling terms to achieve convergence to the target point while maintaining a shape similar to the demonstrated trajectory. With this approach, it is no longer necessary to recompute coefficients for similar trajectories.

\begin{figure*}[!t]
\centering
\subfloat{\includegraphics[scale=0.52]{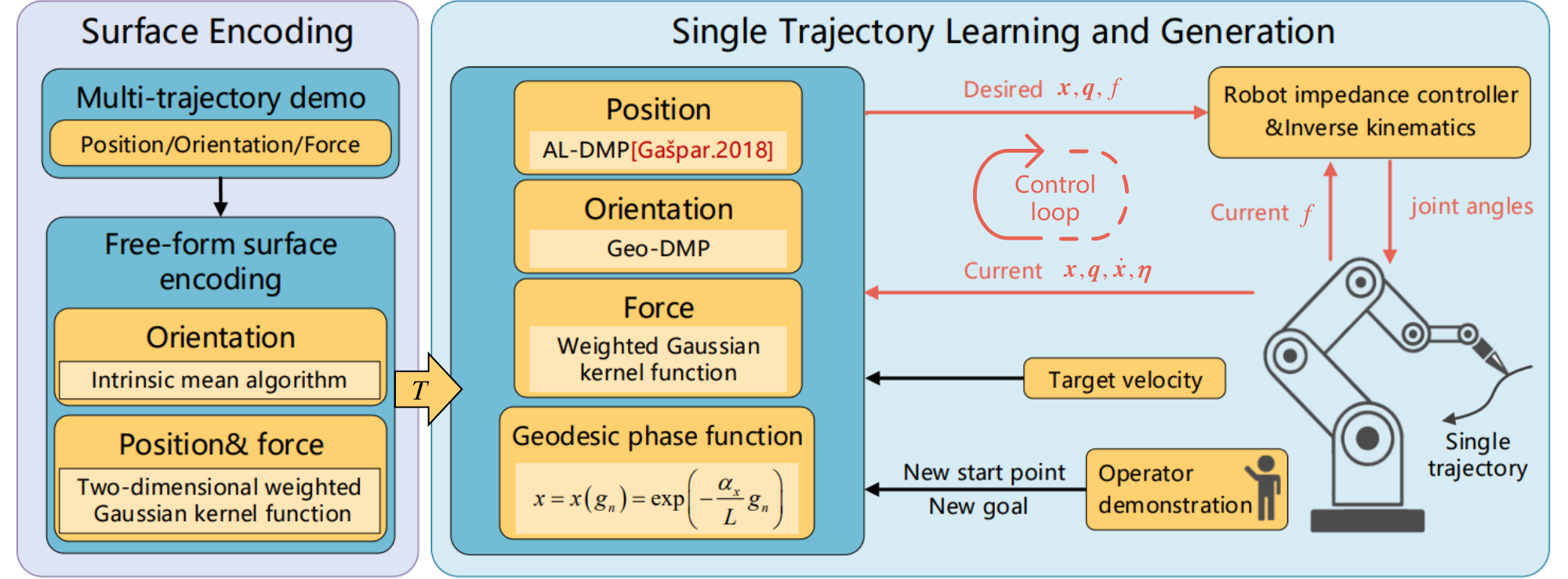}%
}
\hfil
\caption{Robotic grinding skill learning framework. The purple area implements the encoding of surface skills from multiple demonstration trajectories, outputting a position-orientation-force reference trajectory between any two points on the surface. The blue area handles parameter estimation for a single trajectory and real-time generation of robot actions.}
\label{fig2}
\end{figure*}
Free-form surfaces are typically the objects used in grinding tasks. The classical DMPs-based imitation learning framework can only learn similar single trajectories, posing challenges in generating trajectories in scenarios with kinematic constraints, such as on surfaces. Some researchers have attempted to address this issue. Zhou et al. \cite{zhou2022} proposed a combination of Dynamical Movement Primitives (CDMPs), which has the capability to plan combined DMPs on surfaces but requires accurate surface design models. Han et al. \cite{han2022} achieved the generalization of plane demonstration trajectories to surface grinding task trajectories by adding scaling factors and force coupling terms, derived from adaptive impedance control, to the transformation system of original DMPs and optimizing similarity. However, this method relies on force/torque sensors to achieve conformity between the grinding tool and the free-form surface of the workpiece. In addition, the contact state between the grinding tool and the workpiece is complex during the grinding process, making it difficult to apply this method to all conditions. Therefore, encoding surface grinding skills from demonstration trajectories is an urgent problem that needs to be addressed.

Grinding tasks require robots to generate high-precision trajectories, but current research on DMPs still has limitations in this regard, with existing work focusing on trajectory generalization capabilities. Yuan et al. \cite{yuan2019} proposed a method combining reinforcement learning with DMPs for learning in lower limb exoskeleton robots. Lu et al. \cite{lu2021} introduced a DMPs framework tailored to skill generalization constraints and conducted experiments on obstacle avoidance and human-like cooperative operations. Other studies have addressed tasks such as cutter manipulation \cite{straizys2023,wu2021}, grasping \cite{li2017}, and manipulation of deformable objects \cite{cui2022,colome2018}. To address precision issues, Gaspar et al. \cite{gaspar2018} separated the time factor from DMPs and proposed arc-length Dynamic Movement Primitives (AL-DMPs). Building upon this, Wang et al. \cite{wang2023} presented arc-length probabilistic movement primitives (AL-ProMP), which are employed to learn mappings between contact forces and grinding trajectories by taking into account the variability of different grinding tool techniques. However, these efforts have not resolved the issue of encoding orientation in DMPs without time factors. When learning orientation trajectories, it is necessary to consider the orientation constraint of quaternion. If DMPs are used to learn each of the four elements of orientation in $\mathbb{R}^{3}$ space separately \cite{pastor2009}, the generated orientation trajectories would fail to meet the unit norm requirement, resulting in a loss of accuracy. To construct DMPs capable of computing orientation quaternions, Abu-Dakka et al. \cite{abudakka2015} and Ude et al. \cite{ude2014} extended DMPs using the geometric properties of quaternions, projecting orientation quaternions into the tangent space of ${\mathcal{S}}^{3}$ and then performing calculations using metrics in $\mathbb{R}^{3}$ space. However, current orientation encoding methods still implicitly involve time dependence, which leads to difficulties in synchronously generating orientation and positions and significant decreases in accuracy when the speed changes.

Moreover, grinding trajectories require synchronous encoding of position, orientation, and force. Current practices often update the nonlinear forcing term \cite{saveriano2023,liao2022} separately for position, orientation, and force, based on the system's runtime. However, this approach leads to a significant decrease in overall accuracy when external disturbances cause a change in the position of specific DMPs. Furthermore, the feed rate in grinding directly affects the machining quality. Overall scaling or non-uniform adjustment of the feed rate can result in the deformation of learned position-orientation trajectories, causing grinding forces to become misaligned with positions. To establish a synchronous learning framework for force-related skills on free-form surfaces, Gao et al. \cite{gao2019} proposed a method based on Gaussian mixture models to synchronously encode grinding forces and positions on surfaces. While AL-ProMP \cite{wang2023} was used to learn the mapping between contact forces and grinding trajectories, it still does not achieve synchronous encoding of orientation and position. Therefore, a comprehensive synchronous encoding framework is needed to achieve DMPs skill learning in grinding scenarios.

To alleviate the problems of low precision in robotic skills learning and limited generalization capabilities of grinding skills on surfaces, an effective robotic grinding skills learning method is proposed. First, an intrinsic mean algorithm and a two-dimensional weighted Gaussian function are presented to encode the position, orientation, and force of grinding skills on surfaces. Next, the Geo-DMPs based on geodesic length are provided to achieve high-precision encoding of robot orientation with time-independent characteristics. Subsequently, a framework for encoding robotic grinding position, orientation, and force is established. The three variables are synchronized using a phase function based on geodesic length, enabling high-precision control of single grinding skill trajectories. This framework facilitates the generation of robotic grinding actions between any two points on model-free surfaces after demonstration. The entire framework is illustrated in Fig. \ref{fig2}. To sum up, the main contributions of this paper are threefold:

\begin{enumerate}
    \item An encoding method for free-form surfaces is proposed, laying an important foundation for the generalization of DMPs in free-form surface grinding skills. This method achieves the extraction of surface features from multiple grinding demonstration trajectories.

    \item A Geo-DMPs method based on geodesic length is proposed, enabling precise robot orientation encoding. Integrated with AL-DMPs and the weighted Gaussian kernel force encoding method, it constitutes a framework for synchronous encoding of position, orientation, and force, improving trajectory accuracy.

    \item The superiority of the proposed method is validated through comparative experiments on chamfer grinding and free-form surface grinding.
\end{enumerate}

The rest of the paper is organized as follows. Section II describes the encoding method for free-form surfaces. Section III proposes Geo-DMPs for orientation based on the ${\mathcal{S}}^{3}$ manifold geodesic length and the framework for synchronizing position, orientation, and force. Section IV presents the experimental evaluation. Finally, Section V provides the conclusions.

\section{Encoding of Free-form Surfaces}
An important industrial scenario for robotic grinding is free-form surface grinding tasks. However, traditional motion encoding methods do not consider surface geometric constraints and cannot generalize new grinding trajectories on model-free surfaces. Therefore, this section introduces encoding methods for grinding skills on surfaces, aiming to encode free-form surfaces represented by multiple demonstration trajectories. The next section will utilize these encodings to generate a single grinding trajectory. In this section, the two-dimensional weighted Gaussian kernel function for encoding position and force in surface grinding skills is presented in Section II.A. Next, in Section II.B, the intrinsic mean algorithm for encoding orientation in surface grinding skills is proposed. 

\subsection{Surface Position and Force Encoding}
During the demonstration of grinding free-form surfaces, multiple trajectories of the operator's grinding process can be recorded. After alignment using the DTW (Dynamic Time Warping) method, these trajectories can be recorded as a set of point datasets, denoted as
\begin{align}
    \boldsymbol{p}_i=\left\{\boldsymbol{y}_i, \boldsymbol{q}_i, \mathit{f}_i\right\}, \quad i=1,2, \ldots, n,
\end{align}
where \( {{\boldsymbol{y}}_{i}} \in \mathbb{R}^{3} \), \( {{\boldsymbol{q}}_{i}} \in {\mathcal{S}}^{3} \), and \( {{\mathit{f}}_{i}} \in \mathbb{R} \) denote the position, orientation, and grinding force of the \(i\)-th grinding point, respectively. \( {{\boldsymbol{y}}_{i}} = \left[ {{x}_{i}} \ {y}_{i} \ {z}_{i} \right] \) denotes the position of a given grinding point. To facilitate the encoding of skills, it is necessary to parameterize \( {{\boldsymbol{y}}_{i}} \), select the \( \mathrm{XOY} \) plane of the coordinate system as the parameter projection plane, and establish orthogonal \( u \) and \( v \) axes on the plane, such that for each \( {{\boldsymbol{p}}_{i}} \) there is a unique \( (u,v) \) parameter corresponding to it, which can be represented as
\begin{align}
(u_i, v_i) &= \left( s_{x,i} \cdot (x_i - x_{c,i}), s_{y,i} \cdot (y_i - y_{c,i}) \right), 
\end{align}
where $(u_i, v_i)$ represents the projection of \({{\boldsymbol{y}}_{i}}\) onto the \(\mathrm{XOY}\) plane, while \({{x}}_{c,i}\) and \({{y}}_{c,i}\) are the origins of the chosen \(\mathrm{XOY}\) coordinate system. \({{s}}_{x,i}\) and \({{s}}_{y,i}\) are the scaling factors used to adjust the \(x\) and \(y\) scales. Similar to the forcing term in the classical DMPs, free-form surfaces can be encoded using a weighted Gaussian kernel function as
\begin{align}
S(u,v) &= \frac{\sum_{i=1}^{N}\sum_{j=1}^{M} \omega_{ij} \psi_{ij}(u,v)}{\sum_{i=1}^{N}\sum_{j=1}^{M} \psi_{ij}(u,v)} \label{eq:34} ,\\
\psi_{ij}(u,v) &= \exp\left(-h_{ij}\left(\| (u,v) - \boldsymbol{c}_{ij} \|^2\right)\right) .\label{eq:35}
\end{align}

\begin{figure}[h]
    \centering
    \includegraphics[scale=0.11]{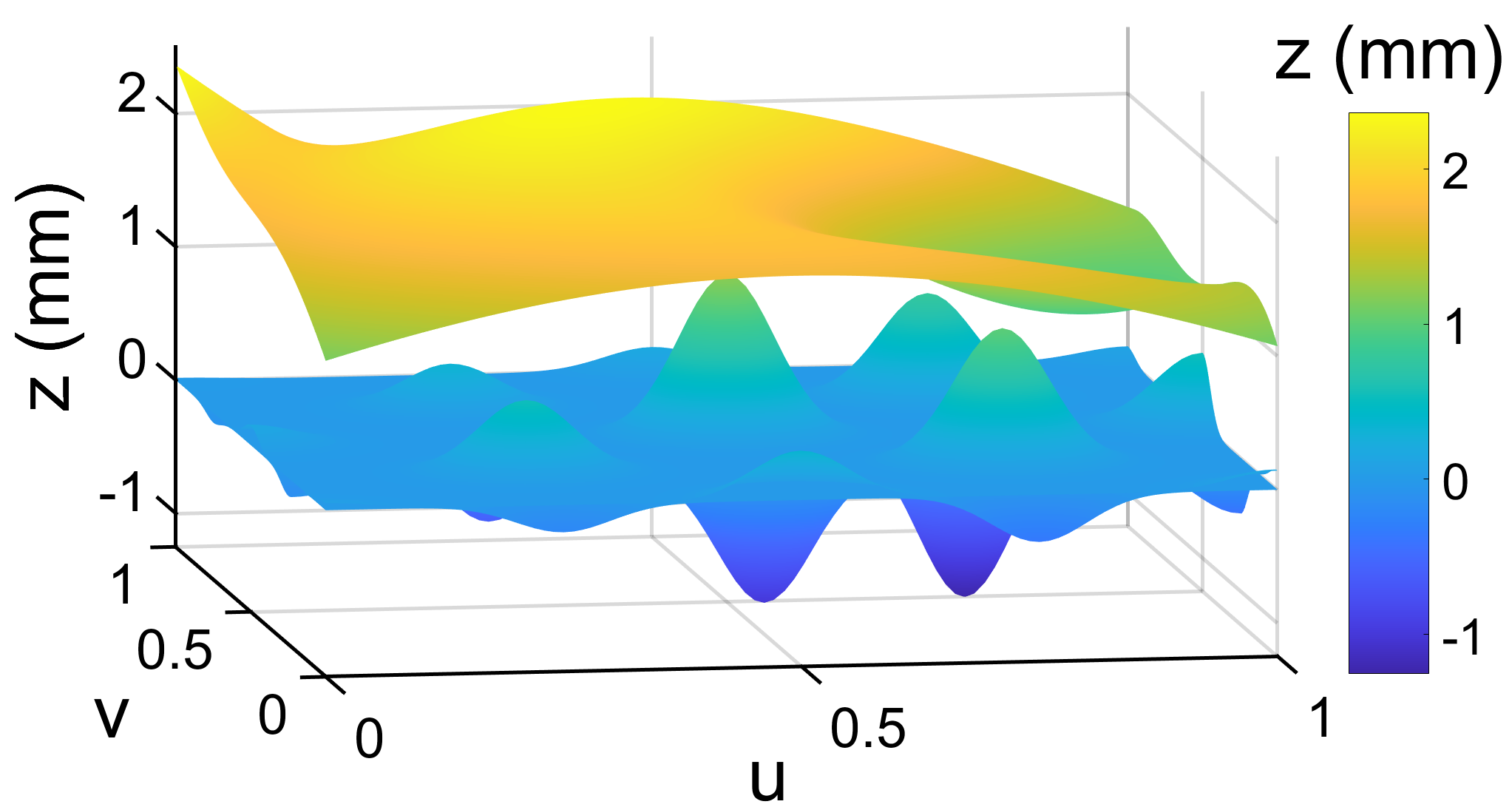}
    \caption{Encoding surfaces with two-dimensional weighted Gaussian kernel function. The upper surface represents the free-form surface obtained from encoding, while the lower surface corresponds to the weighted Gaussian kernel functions.}
    \label{fig5}
\end{figure}

${\psi}_{ij}(u,v)$ denotes the two-dimensional Gaussian kernel function. Define the center $\boldsymbol{c}_{ij}$ of the Gaussian kernel function with a uniform distribution through the hyperparameter method. The weight coefficient $\omega_{ij}$ can be estimated using the method of Locally Weighted Regression, as shown in Fig. \ref{fig5}. For a particular point \({{\boldsymbol{y}}_{i}}\), there is only one corresponding grinding force $f_i$, which can be encoded using the same method.
    
\subsection{Intrinsic Mean Algorithm for Orientation Manifold}
 For a randomly chosen position point ${\boldsymbol{y}_{D}}$ on the surface, the corresponding orientation ${\boldsymbol{q}_{D}}$ can be estimated by the weighted mean of the orientations of $K$ teaching points ${\boldsymbol{y}_{i}}$ whose distances from ${\boldsymbol{y}_{D}}$ do not exceed $d$.Here, $d$ represents the upper limit of the distance between sample points and ${\boldsymbol{y}_{D}}$.

Cluster center estimation methods usually involve minimizing the sum of squares of distances from the data points to their cluster centers, i.e. $\underset{{\boldsymbol{q}_{D}}}{\arg \min } \sum\limits_{i=1}^{K} \|{\boldsymbol{q}_{i}}-{{\boldsymbol{q}_{D}}}\|^2$. However, the orientation data do not belong to the Euclidean space, and calculating the mean of the orientation quaternion directly in the Euclidean space will result in reduced accuracy. Therefore, based on the definition of the distance metric on the ${{\mathcal{S}}^{3}}$ sphere \cite{Ude1999}, an objective function for estimating the cluster centers in the streamlined space can be represented as
\begin{align}
\underset{\boldsymbol{q}_{D}}{\arg \min }\sum\limits_{i=1}^{K} \|2\log^{\boldsymbol{q}}{({\boldsymbol{q}_{i}}*\overline{\boldsymbol{q}}_{D})}\|^2.
\end{align}
where $*$ is quaternion multiplication and can be defined as
\begin{align}
    {{\boldsymbol{q}}_1*{\boldsymbol{q}}_2} &= {\left(\nu_1+{\boldsymbol{u}}_1\right)*\left(\nu_2+{\boldsymbol{u}}_2\right)}\notag\\
    &= {\left(\nu_1\nu_2-{\boldsymbol{u}}_1^T*{\boldsymbol{u}}_2\right)} + {\left(\nu_1{\boldsymbol{u}}_2+\nu_2{\boldsymbol{u}}_1+{\boldsymbol{u}}_1\times{\boldsymbol{u}}_2\right)}.
\end{align}

\begin{figure}[H]
    \centering
    \includegraphics[scale=0.1]{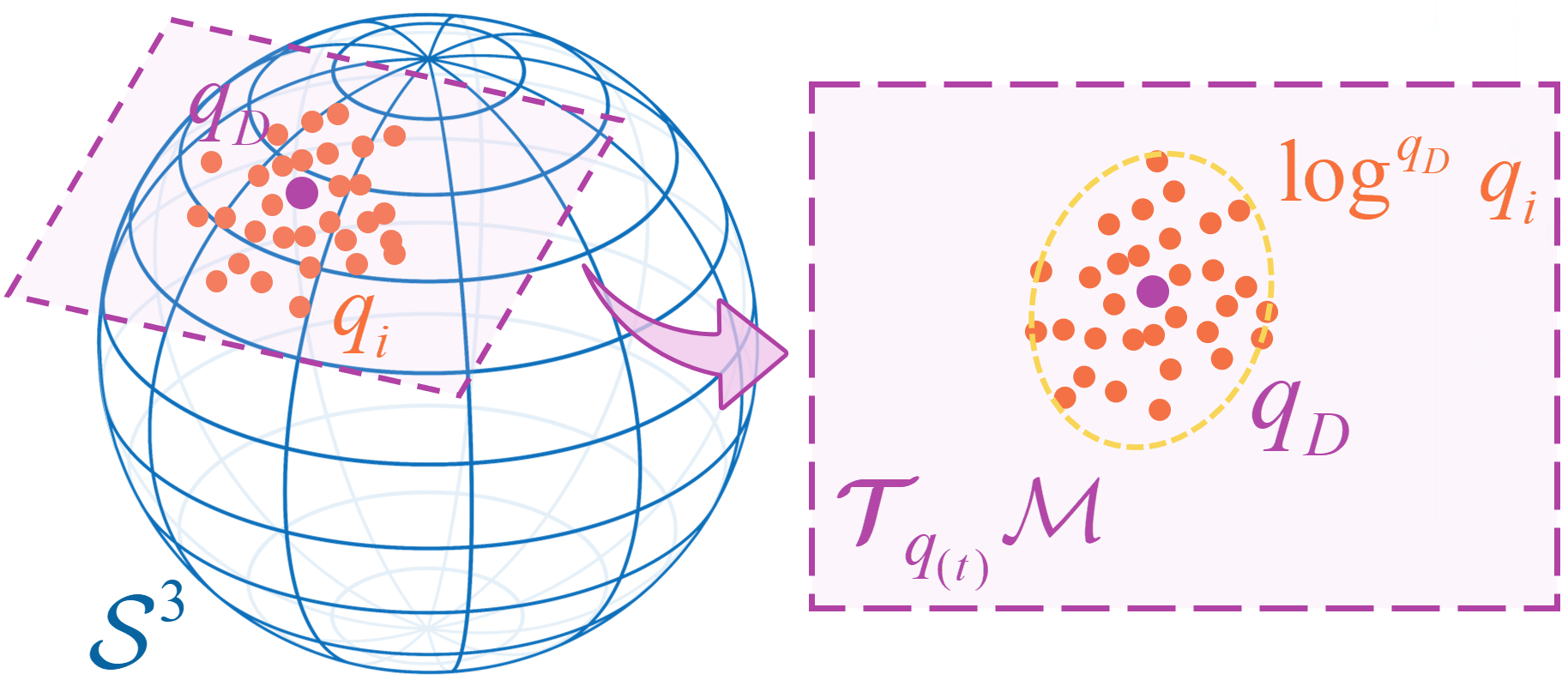}
    \caption{Intrinsic cluster center algorithm on the manifold illustration.}
    \label{fig6}
\end{figure}

The intrinsic cluster center algorithm on the manifold illustration is shown in Fig. \ref{fig6}. The purple plane represents the tangent space $ \mathcal{T}_{q(t)}\mathcal{M}$  at any point $ q_D $ on the manifold. In this space, Euclidean space metrics can be used to perform linear combination operations. This method effectively estimates the average orientation within the selected region, delivering more accurate results when the demonstration data are uniformly and densely distributed. However, demonstration points tend to be randomly dispersed in scenarios where a human operator demonstrates surface grinding skills. When the locations of demonstration points are unevenly distributed, (5) struggles to accurately estimate the orientation ${\boldsymbol{q}_{D}}$ at position ${\boldsymbol{y}_{D}}$ on the free-form surface. Therefore, it is necessary to introduce distance weights for the demonstration point locations, denoted as $\left(d-\left\| {\boldsymbol{y}_{D}}-{\boldsymbol{y}_{i}} \right\|\right)/d$. Consequently, based on the distance metric on the ${{\mathcal{S}}^{3}}$ sphere and the distance weights associated with the demonstration points, the optimized objective function can be formulated as
\begin{align}
\underset{{\boldsymbol{q}_{D}}}{\arg \min }\sum\limits_{i=1}^{K}\frac{d-\left\| {\boldsymbol{y}_{D}}-{\boldsymbol{y}_{i}} \right\|}{d} \cdot \left\|2\log^{\boldsymbol{q}}({{\boldsymbol{q}}_{i}}*\overline{\boldsymbol{q}}_{D})\right\|^2.
\end{align}

\section{Grinding Skills Learning and Single Trajectory Generation}

This section presents the method for generalizing grinding skill trajectories between any two points on a surface, starting from the encoding introduced in the previous section. First, in Section III.A, the classical DMPs are introduced, and position trajectories are generated based on AL-DMPs \cite{gaspar2018}. Next, in Section III.B, the geodesic length-based orientation Geo-DMPs on the ${S}^3$ manifold are proposed for generating orientation trajectories. Finally, in Section III.C, the encoding method for force is introduced, along with the method for controlling the robot to generate grinding actions based on the synchronization encoding framework. The illustration is shown in Fig. \ref{fig2}.

\subsection{Arc-length DMPs for Position }
\subsubsection{Classical DMPs}
The classical DMPs consist of a set of nonlinear differential equations \cite{ijspeert2013}, which can be used to learn the mapping from position $y(t)\in \mathbb{R}^{3}$ and velocity $\dot{y}$ to acceleration $\ddot{y}$ from demonstrated trajectories while ensuring convergence to the goal position. Their expression is
\begin{align}
    &\left\{
    \begin{aligned}
        &\tau\dot{\boldsymbol{z}}=\alpha_{z}\left(\beta_{z}(\boldsymbol{g}-\boldsymbol{y})-\boldsymbol{z}\right)+\boldsymbol{f}(x) \\
        &\tau\dot{\boldsymbol{y}}=\boldsymbol{z} \\
        &\tau\dot{x}=\alpha_{x}x
    \end{aligned},
    \right.
\end{align}
where $\boldsymbol{z}\in \mathbb{R}^3$ as an auxiliary variable denotes the scaled velocity, and $\boldsymbol{g}\in \mathbb{R}^3$ denotes the target position of the trajectory. The parameters $\alpha_z$ and $\beta_z$ define the behavior of the second-order system. $\boldsymbol{f}(x)$ is a forcing term, where $x\in \mathbb{R}$ represents the time phase variable, and $\alpha_{x}$ corresponds to the coefficients of the regularization system.

DMPs generalize trajectories separately in the $x$, $y$, and $z$ directions, aligning the phase of each degree of freedom through time series alignment. However, variations in grinding speed can lead to changes in trajectory shape. This is because DMPs rely on the velocity and shape of a movement but cannot separate the time and space of an action. AL-DMPs \cite{gaspar2018}, instead, can effectively separate the time component,  ensuring the accuracy of the learned action trajectories.

\subsubsection{Arc-length DMPs}
The arc length of the trajectory can be denoted as
\begin{align}
    s(t)=\int_0^t\|\dot{\boldsymbol{y}}(u)\|du.
\end{align}

Thus, the total arc length of the trajectory moving in time period $[0, T]$ can be denoted as
\begin{align}
    L=\int_0^T\|\dot{\boldsymbol{y}}(t)\|dt.
\end{align}

Based on (8) and (9), the arc-length derivative of position and velocity in classical DMPs yields the AL-DMPs (arc-length DMPs) \cite{gaspar2018}, which can be represented as
\begin{align}
    \begin{cases}
    L{\boldsymbol{z}}' &= \alpha_z\big(\beta_z({\boldsymbol{g}}-{\boldsymbol{y}})-{\boldsymbol{z}}\big)+{\boldsymbol{F}}(x)\\
    L{\boldsymbol{y}}' &= \boldsymbol{z}\\
    L{x}' &= -\alpha_x{x}
    \end{cases},
\end{align}
where the forcing term $\boldsymbol{F}(x)$ is defined as
\begin{equation}
\begin{aligned}
\boldsymbol{F}(x) & =\operatorname{diag}\left(\boldsymbol{g}-\boldsymbol{y}^0\right) \frac{\sum_{i=1}^N \boldsymbol{w}_i \psi_i(x)}{\sum_{i=1}^N \psi_i(x)} x, 
\end{aligned}
\end{equation}
\begin{equation}
\begin{aligned}
\psi_i(x) & =\exp \left(-h_i\left(x-c_i\right)^2\right)
\end{aligned},
\end{equation}
where $\operatorname{diag}\left(\boldsymbol{g}-\boldsymbol{y}^0\right)\in \mathbb{R}^{3\times3}$ denotes a diagonal matrix, $\boldsymbol{y}^0$ denotes the start position, $\boldsymbol{g}$ denotes the end position of the teachable data, and $\psi_i(x)$ denotes a Gaussian basis function with center position $c_i$ and width $h_i$. AL-DMPs differ from the classical DMPs because the derivatives are all in arc lengths rather than time. 

For any two points ${\boldsymbol{y}_{A}}$ and ${\boldsymbol{y}_{B}}$ selected as the start and end points of the desired trajectory on the surface, the $u,v$ parameters of each interpolated point on the trajectory can be derived from
\begin{equation}
\begin{aligned}
    {\boldsymbol{y}_{D}}(u,v) &= \left( 1-{{t}_{u}} \right){\boldsymbol{y}_{A}}(u,v) + {{t}_{u}}{\boldsymbol{y}_{B}}(u,v), \\
    {{t}_{u}} &= \frac{i}{{{N}_{u}}}, \quad i=0,1,\ldots, {{N}_{u}},
\end{aligned}
\end{equation}
where ${\boldsymbol{y}_{D}}(u, v)$ denotes the $u,v$ parameter corresponding to the interpolation point on the generated reference trajectory, and ${{t}_{u}}\in (0,1)$ is the interpolation parameter determined by the number of interpolation points ${{N}_{u}}$. Based on the generated $u,v$-interpolated sequence, the coordinates ${\boldsymbol{y}_{D}}({{u}_{i}},{{v}_{i}})=S({{u}_{i}},{{v}_{i}})$ of each interpolated point can be calculated with the weighted 2D Gaussian kernel function described in (3). The weights $\boldsymbol{w}_{i}$ can be estimated by
\begin{equation}
\begin{aligned}
& \sum_{i=1}^N \frac{\Psi_i\left(x\right)}{\sum_{j=1}^N \Psi_j\left(x\right)} \boldsymbol{w}_i = \\
& \frac{1}{x} \operatorname{diag}\left(\boldsymbol{g}-\boldsymbol{y}^0\right)^{-1}\left(L^2 \boldsymbol{y}_D^{\prime \prime}-\alpha_z\left(\beta_z\left(\boldsymbol{g}-\boldsymbol{y}_D\right)-L \boldsymbol{y}_D^{\prime}\right)\right).
\end{aligned}
\end{equation}

In AL-DMPs, the phrase is only correlated with arc length, meaning that the shape of the generated robot's motion remains unchanged regardless of velocity. However, due to the gimbal lock problem associated with Euler angles and quaternion constraints, it may be challenging to apply this approach directly to the robot's orientation control.

\subsection{Geodesic-length DMPs for Orientation}
\subsubsection{Estimation of Geodesic Length}
Orientation belongs to non-Euclidean space. Traditional Euclidean space algorithms encounter issues with metric calculation and closure when processing demonstration data and lack statistical analysis of the manifold characteristics. Therefore, it is necessary to establish a metric for orientation data.

For any two points $\boldsymbol{A}$ and $\boldsymbol{B}$ on the manifold $\mathcal{M}$, define $\gamma(t)$ as a continuous differentiable curve connecting $\boldsymbol{A}$ and $\boldsymbol{B}\in \mathcal{M}$, with length given as \cite{abu2022}
\begin{align}
    \mathcal{L}_{\boldsymbol{A}}^{\boldsymbol{B}}(\gamma)=\int_0^1<\gamma(t),\gamma(t)>dt.
\end{align}

The local shortest path connecting two points is defined as the geodesic, and the length of the geodesic is defined as the Riemannian distance between the two points
\begin{align}
    \mbox{dist}({\boldsymbol{A}},{\boldsymbol{B}})=\min \mathcal{L}_{\boldsymbol{A}}^{\boldsymbol{B}}(\gamma).
\end{align}

For any two orientations, direct vector operations between them are infeasible. However, for any point on the manifold $\mathcal{M}$, its tangent space $\mathcal{T}_{\boldsymbol{p}\mathcal{M}}$ is a vector space composed of all tangent vectors at that point, and vector operations within the tangent space adhere to vector algebra principles.

Logarithmic mapping and exponential mapping are used to map a point on the manifold to the tangent space and map a point in the tangent space back to the manifold, respectively:
\begin{align}
    &\log^{\boldsymbol{p}}(\cdot):\mathcal{M}\mapsto {\mathcal{T}_{\boldsymbol{p}}\mathcal{M}},\\
    &\exp^{\boldsymbol{p}}(\cdot):{\mathcal{T}_{\boldsymbol{p}\mathcal{M}}}\mapsto \mathcal{M}.
\end{align}

\begin{figure}[t]
    \centering
    \includegraphics[scale=0.12]{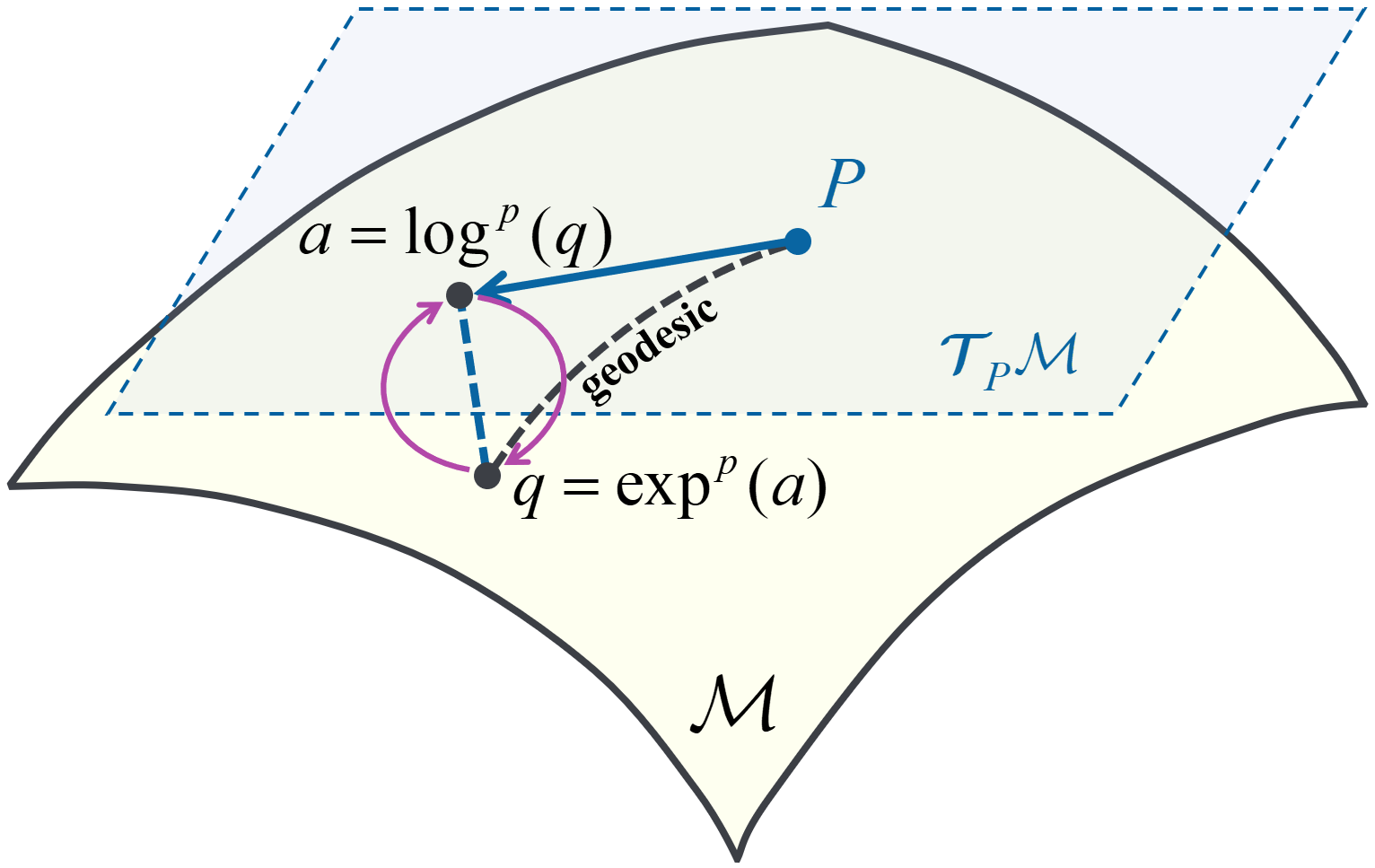}
    \caption{The purple arrow represents the logarithmic and exponential mappings between the manifold $\mathcal{M}$ and the tangent space $\mathcal{T}_{\boldsymbol{p}}\mathcal{M}$. For a point $\boldsymbol{q}$ on the manifold needs to be mapped to the tangent space $\mathcal{T}_{\boldsymbol{p}}\mathcal{M}$  in order to perform linear operations.}
    \label{fig3}
\end{figure}

Unit quaternion $\boldsymbol{q}=v+\boldsymbol{u}\in \mathcal{S}^3$ can be used to represent the poses of motion trajectories, where $v\in \mathbb{R}$ and $\boldsymbol{u}\in \mathbb{R}^3$ correspond to the real and imaginary parts of the quaternion, respectively. As shown in Fig. \ref{fig3}, the logarithmic mapping and exponential mapping can be defined as \cite{Ude1999}
\begin{align}
    \log^{\boldsymbol{q}}({\boldsymbol{q}})=\begin{cases}
    \arccos(\nu)\frac{{\boldsymbol{u}}}{\|{\boldsymbol{u}}\|},\quad{\boldsymbol{u}}\neq0\\
    \begin{bmatrix}0&0&0\end{bmatrix}^T,\quad\text{otherwise}\end{cases},
\end{align}
\begin{align}
    \exp({\boldsymbol{a}})=\begin{cases}\cos(\|\boldsymbol{a}\|)+\sin(\|{\boldsymbol{a}}\|)\frac{{\boldsymbol{a}}}{\|{\boldsymbol{a}}\|},{\boldsymbol{a}}\neq\boldsymbol{0}\\
    1+\begin{bmatrix}0&0&0\end{bmatrix}^T,\text{otherwise}\end{cases}.
\end{align}

Based on (19), the distance between two orientation quaternions on the $\mathcal{S}^3$ sphere can be defined as
\begin{align}
    d({\boldsymbol{q}}_{1},{\boldsymbol{q}}_{2})=\begin{cases}
        2\pi,\quad \boldsymbol{q}_1*\bar{\boldsymbol{q}}_2=-1+\begin{bmatrix}
            0&0&0
        \end{bmatrix}\\
        2\|{\log^{\boldsymbol{q}}(\boldsymbol{q}_1*\bar{\boldsymbol{q}}_2)}\|,\quad\text{otherwise}
    \end{cases},
\end{align}

Based on (21), a reference trajectory of orientation obtained from section II.B or a demonstration can be represented as
\begin{align}
    T_{demo}=\{\boldsymbol{q}_k\}_{k=1}^K,\boldsymbol{q}_k\in \mathcal{S}^3,
\end{align}
and the total length of the geodesic line for its orientation change can be expressed as
\begin{align}
    L=\sum_{k=1}^Kd(\boldsymbol{q}_k,\boldsymbol{q}_{k+1}).
\end{align}

The geodesic length of the trajectory at time $n$ can be defined as
\begin{align}
    g=\begin{cases}
        \sum_{k=1}^n2\|{\log^{\boldsymbol{q}}(\boldsymbol{q}_k*\bar{\boldsymbol{q}}_{k+1})}\|,\quad k\ge2\\
        0,\quad k=1
    \end{cases}.
\end{align}
\subsubsection{Expression of Geodesic-length DMPs for Orientation}
To establish a time-independent mapping from orientation and angular velocity to angular acceleration, we propose the geodesic length DMPs (Geo-DMPs) as
\begin{equation}
    \begin{aligned}
        \begin{cases}
        L\boldsymbol{\eta}'=\alpha_z(\beta_z 2 \log^{\boldsymbol{q}}(\boldsymbol{q}_g*\bar{\boldsymbol{q}})-\boldsymbol{\eta})+\boldsymbol{f}_{\boldsymbol{q}}(x)\\
        L\boldsymbol{q}'=\frac{1}{2}\boldsymbol{\eta}*\boldsymbol{q}\\
        Lx'=-\alpha_xx
        \end{cases},
    \end{aligned}
\end{equation}
where $\boldsymbol{q}_g$ represents the target orientation, and $\boldsymbol{q}$ is the current orientation. $L$ represents the total geodesic length of the motion, which is calculated using (24). $\boldsymbol{\eta}\in \mathbb{R}^3$ is the result of differentiating orientation variation with respect to geodesic length. In the calculations, $\boldsymbol{\eta}$ is considered as a quaternion with a zero scalar part. $\boldsymbol{\eta}'$ represents the derivative of $\boldsymbol{\eta}$ with respect to the geodesic length. Given the initial condition $x(0)=1$, $x$ can be analytically solved through
\begin{align}
    x=\exp\left(-\frac{\alpha_x}{L}g\right).
\end{align}

\begin{figure}[t]
    \centering
    \includegraphics[scale=0.13]{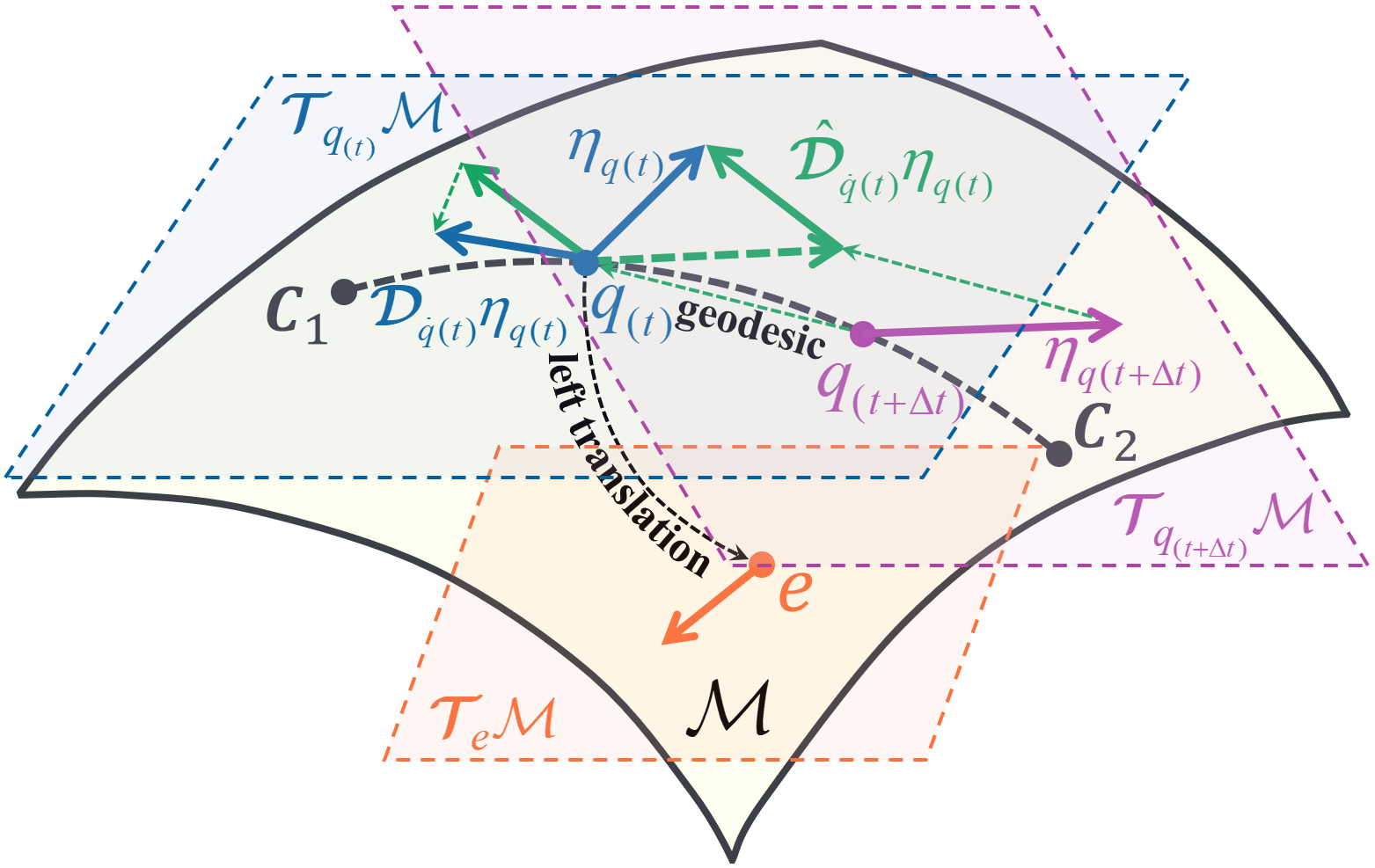}
    \caption{The connection between the tangent spaces of $\boldsymbol{q}_{(t)}$ and $\boldsymbol{q}_{(t+\Delta t)}$ maps the vector $\eta_{\boldsymbol{q}(t+\Delta t)}$ in the tangent space $\mathcal{T}_{\boldsymbol{q}_{(t+\Delta t)}}\mathcal{M}$ to the tangent space $\mathcal{T}_{\boldsymbol{q}_{(t)}}\mathcal{M}$ of $\boldsymbol{q}_{(t)}$.}
    
    \label{fig4}
\end{figure}

To compute the geodesic length derivative of the quaternion, we first provide its first-order time derivative
\begin{align}
    \dot{\boldsymbol{q}}=&(\dot{q}_1,\dot{q}_2,\dot{q}_3,\dot{q}_4)\notag\\
    =&\frac{1}{2}\dot{g}\boldsymbol{\eta}*\boldsymbol{q},
\end{align}
where $\boldsymbol{\eta}$ represents the derivative of angular velocity concerning arc length, and $\dot{g}$ represents the derivative of the geodesic length with respect to the rotation angle. Based on the form of the Connection's derivation, $\boldsymbol{\omega}$ and $\dot{\boldsymbol{\omega}}$ can be expressed as
\begin{align}
    \begin{cases}
        \boldsymbol{\omega}=\boldsymbol{\eta} \dot{g}\\
        \dot{\boldsymbol{\omega}}=\mathcal{D}_{\dot{\boldsymbol{q}}(t)}\boldsymbol{\eta} \dot{g}_{\boldsymbol{q}(t)}=\mathcal{X}_{\dot{\boldsymbol{q}}(t)}\dot{g}\boldsymbol{\eta}_{\boldsymbol{q}(t)}+\dot{g}\cdot\mathcal{D}_{\dot{\boldsymbol{q}}(t)}\boldsymbol{\eta}_{\boldsymbol{q}(t)}
    \end{cases},
\end{align}
where ${\mathcal{X}_{\dot{\boldsymbol{q}}(t)}}$ is an operator, representing the directional derivative of $\dot{g}\boldsymbol{\eta}_{\boldsymbol{q}(t)}$ in the direction of $\dot{q}(t)$.
 $\mathcal{D}_{\dot{\boldsymbol{q}}(t)}\boldsymbol{\eta} \dot{g}_{\boldsymbol{q}(t)}$ is the directional derivative with respect to the vector field $\boldsymbol{\eta} \dot{g}$ at $\boldsymbol{q}$ along the $\dot{\boldsymbol{q}}$ direction. $\boldsymbol{\omega}$ denotes the rotation quantity defined in the tangent space $\mathcal{T}_{\boldsymbol{e}\boldsymbol{q}}$ at the unit element $\boldsymbol{q}_0(\boldsymbol{q}_0=[1,0,0,0])$. $\boldsymbol{\omega}$ is the left-invariant vector field over the entire manifold, generated by the action of the tangent space vector at the identity element through elements $\boldsymbol{q}$ in the manifold. $\dot{\boldsymbol{q}}$ represents the rotational quantity at $\mathcal{T}_{\boldsymbol{q}(t)}\boldsymbol{q}$, which can be obtained by left translation of the rotational quantity in $\mathcal{T}_{\boldsymbol{e} \boldsymbol{q}}$ under the action of $\boldsymbol{q}$. Fig. \ref{fig4} shows the connection between tangent spaces on a manifold.

According to the definition of the ambient space, the relationship between $\boldsymbol{q}$ as a submanifold of $\mathbb{R}^4$ and the connection of $\mathbb{R}^4$ is established as
\begin{align}
    \mathcal{D}_{\dot{\boldsymbol{q}}(t)}\boldsymbol{\eta}_{\boldsymbol{q}(t)}=&\bar{\boldsymbol{q}}(t)\cdot(\hat{\mathcal{D}}_{\dot{\boldsymbol{q}}(t)}\boldsymbol{\eta}_{\boldsymbol{q}(t)}\notag\\
    &-<\hat{D}_{\dot{\boldsymbol{q}}(t)}\boldsymbol{\eta}_{\boldsymbol{q}(t)},\boldsymbol{q}(t)>\boldsymbol{q}(t))\in \mathcal{T}_{\boldsymbol{q}(t)}\boldsymbol{q},\\
    \boldsymbol{\eta}'=&\Bar{\boldsymbol{q}}(t)*\mathcal{D}_{\dot{\boldsymbol{q}}(t)}\boldsymbol{\eta}_{\boldsymbol{q}(t)},\\
    \hat{\mathcal{D}}_{\dot{\boldsymbol{q}}(t)}\boldsymbol{\eta}_{\boldsymbol{q}(t)}=&\lim_{\Delta t\to0}(\frac{\dot{q}_1(t+\Delta t)-\dot{q}_1(t)}{\dot{g}(t)},\frac{\dot{q}_2(t+\Delta t)-\dot{q}_2(t)}{\dot{g}(t)},\notag\\
    &\frac{\dot{q}_3(t+\Delta t)-\dot{q}_3(t)}{\dot{g}(t)},\frac{\dot{q}_4(t+\Delta t)-\dot{q}_4(t)}{\dot{g}(t)}).
\end{align}

The analytical expressions for $\boldsymbol{\eta}$ and $\boldsymbol{\eta}'$ obtained from the above equations are
\begin{equation}
    \begin{aligned}
        \boldsymbol{\eta}=&\frac{\boldsymbol{\omega}}{\dot{g}},\\
        \boldsymbol{\eta}'=&\bar{\boldsymbol{q}}(t)*\frac{\dot{\boldsymbol{\omega}}\dot{g}-\boldsymbol{\omega} \mathcal{X}_{\dot{\boldsymbol{q}}(t)}\dot{g}}{\dot{g}^2}.
    \end{aligned}
\end{equation}

And
\begin{align}
    \dot{x}=\frac{d}{dt}x(g(t))=x'\dot{g}.
\end{align}

The forcing term $\boldsymbol{f}_{\boldsymbol{q}}(x)$ is formulated using a set of Gaussian basis functions as
\begin{align}
    {\boldsymbol{f}_{\boldsymbol{q}}(x)}=\operatorname{diag}(\log^{\boldsymbol{q}}(\boldsymbol{q}_{g}*\Bar{\boldsymbol{q}}_0))\frac{\sum_{i=1}^{N}\boldsymbol{w}_i^o\psi_i(x)}{\sum_i^N\psi_i(x)}x,
\end{align}
where $\operatorname{diag}(\log^{\boldsymbol{q}}(\boldsymbol{q}_{g}*\Bar{\boldsymbol{q}}_0))\in \mathbb{R}^{3\times3}$ is a diagonal matrix, and $\boldsymbol{q}_0$ is the initial state of the demonstration data, $\boldsymbol{q}_g$ is the final state, and $N$ is the number of Gaussian basis functions adjusted by hyperparameters.

$\boldsymbol{w}_i^o\in \mathbb{R}^3$ in (28) can be obtained by solving the linear system
\begin{align}
\frac{\sum_{i=1}^{N}\boldsymbol{w}_i^o\psi_i(x)}{\sum_i^N\psi_i(x)}x=&\operatorname{diag}(\log^{\boldsymbol{q}}(\boldsymbol{q}_{g}*\Bar{\boldsymbol{q}}_0))^{-1}(\tau\dot{\boldsymbol{\eta}}\notag\\
&-\alpha_z(\beta_z(2\log^{\boldsymbol{q}}(\boldsymbol{q}_g*\Bar{\boldsymbol{q}}_k)-\boldsymbol{\eta}))),
\end{align}
where the phase $x$ can be estimated by (27).

Normally, the robot controller will control the robot according to a fixed time period, but the output of Geo-DMPs is the second-order geodesic length derivative of the orientation and therefore cannot be used directly for robot control. The velocity and acceleration of the robot's orientation in the time dimension can be calculated using (29). Then, the expectation of the robot's orientation at the next instant can be obtained using the logarithmic mapping
\begin{align}
    \boldsymbol{q}(t+\Delta t)=\exp(\frac{\Delta t}{2}\frac{\boldsymbol{\eta}(t)}{L})*\boldsymbol{q}(t).
\end{align}

\subsection{Grinding Force Learning and Motion Generation}
In robotic grinding tasks, the grinding force directly influences the material removal rate, requiring varying normal grinding forces for workpiece areas with different removal rate requirements. When robots are used to perform grinding tasks, a hybrid force-orientation control method is commonly employed to decouple the pose and force into independent subspaces. Tangential and rotational directions are controlled for the position, while an impedance controller is utilized for force control in the normal direction along the trajectory or surface. Therefore, it is necessary to encode the normal force corresponding to each pose in the grinding process. Force-related variables in grinding tasks do not need to converge from an initial value to a final value, making a PD controller in DMPs physically meaningless. Therefore, the force during the grinding process can be encoded by the weighted Gaussian kernel function as
\begin{align}
    f=\frac{\sum_{i=1}^N w_{f_i}\psi_{f_i}(x)}{\sum_{i=1}^N\psi_{f_i}(x)},
\end{align}
where $\psi_{f_i}(x)$ is a radial basis function, and its center and variance can be determined according to the number of basis functions. The trajectories generated by DMPs can be used as reference trajectories for the impedance controller. The grinding normal force can be obtained by regression based on the geodesic length that the current robot has moved during the online execution, enabling high synchronization between the grinding normal force and the robot's orientation.

To generate actions based on learned skills, the next desired position and force need to be updated in the control loop, as shown in Fig. \ref{fig2}. The next desired position $\boldsymbol{x}$ and orientation $\boldsymbol{q}$ are computed and generated by AL-DMPs and Geo-DMPs, respectively. (38) is used to calculate the next desired force. Subsequently, the robot's desired joint angles are obtained through the inverse kinematics and the impedance controller. During this process, the robot provides feedback of its current position $\boldsymbol{x}$, velocity $\dot{\boldsymbol{x}}$, orientation $\boldsymbol{q}$, and angular velocity $\boldsymbol{\eta}$ to AL-DMPs and Geo-DMPs, respectively.

\section{Experiment Evaluation}
In this section, several experiments were designed to validate the proposed grinding skills learning method. The experiments were conducted for two typical industrial scenarios: chamfer grinding and free-form surface grinding.
\subsection{Chamfer Grinding}
\subsubsection{System configuration}
We designed an experiment to grind the edges of a C-shaped workpiece. This scenario requires high accuracy in the robot's orientation and precise synchronization of position, orientation, and force.
\begin{figure}[h]
    \centering
    \includegraphics[scale=0.24]{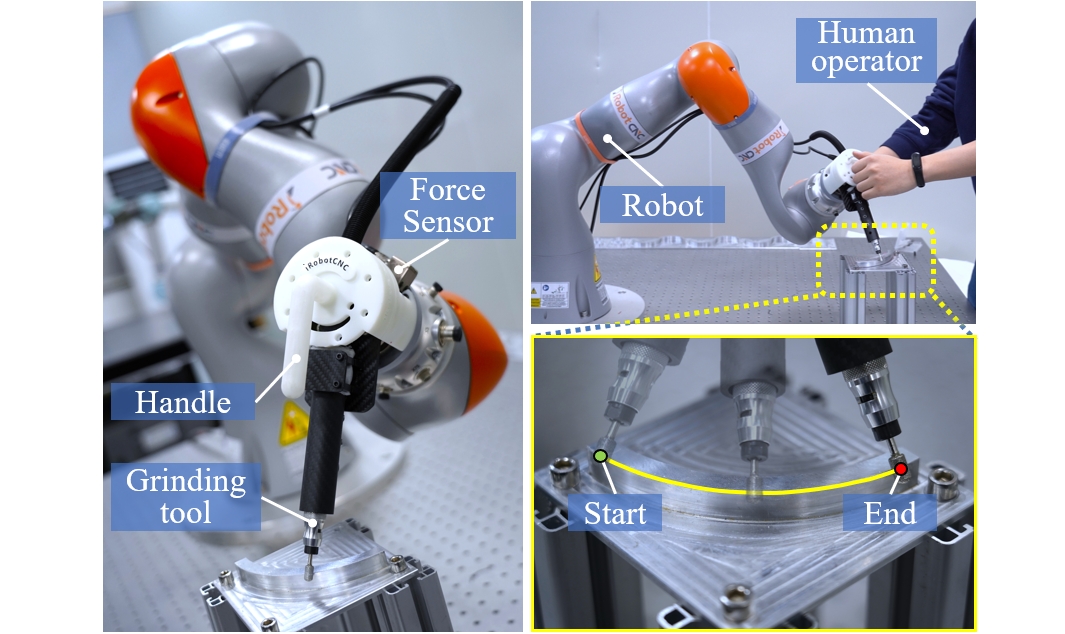}
    \caption{Experimental setup for a human operator demonstrating chamfer grinding to a robot, showing the robotic platform and the chamfering trajectory on a C-shaped workpiece.}
    \label{fig7}
\end{figure}

The experimental setup is shown in Fig. \ref{fig7}. The platform consists of a KUKA iiwa 14 R820 robot, ATI force sensors, a pneumatic grinding pen, and a drag handle. The pneumatic grinding pen is connected to the robot's end flange via the ATI force sensors, and the drag handle is mounted directly on the robot's end flange. The human operator can demonstrate the grinding path and normal force to the robot by dragging. The force sensors can directly capture the grinding force applied to the pneumatic grinding pen. The control frequencies for DMPs and the impedance controller were set to 1000 Hz and 125 Hz, respectively.

\subsubsection{Geo-DMPs Experiment}
To verify the encoding accuracy of Geo-DMPs, we conducted an experiment on orientation trajectory learning and generation on a C-shaped workpiece. In this experiment, we compared Geo-DMPs with the widely recognized most effective Quat-DMPs \cite{saveriano2023,abudakka2015} and the optimized Riemannian-DMP \cite{liao2022} based on it. The effectiveness of DMPs in completing grinding tasks depends on the quality of demonstration by human operators. Therefore, the operator practiced dragging the robot to grind the edges of the C-shaped workpiece 200 times before the demonstration. During the demonstration, the pneumatic grinding tool was not activated, and the robot recorded the end-effector trajectory while the force sensor recorded the normal grinding force. Fig.8 shows the demonstrated trajectory of the workpiece. We encoded the position, orientation, and force separately using the proposed method in Section III. Then, the encoded data were input into the Geo-DMPs to reproduce the demonstration start to end points, as shown in Fig. 8.

\begin{figure}[h]
    \centering
    \includegraphics[scale=0.25, angle=0, origin=c]{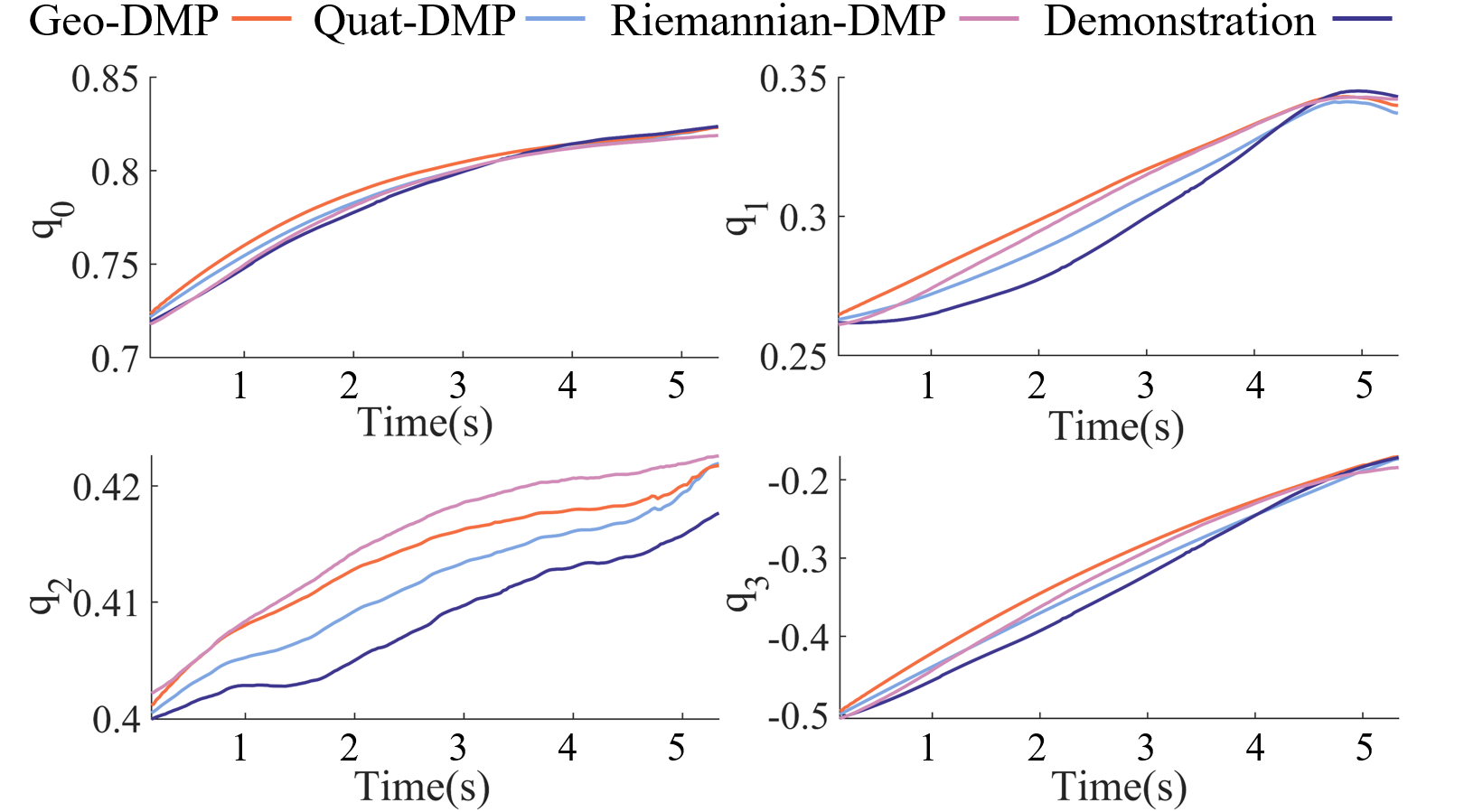}
    \caption{Robot's orientation quaternions for demonstration and experiment.}
    \label{fig8}
\end{figure}

\begin{figure}[t]
    \centering
    \includegraphics[scale=0.2, angle=0, origin=c]{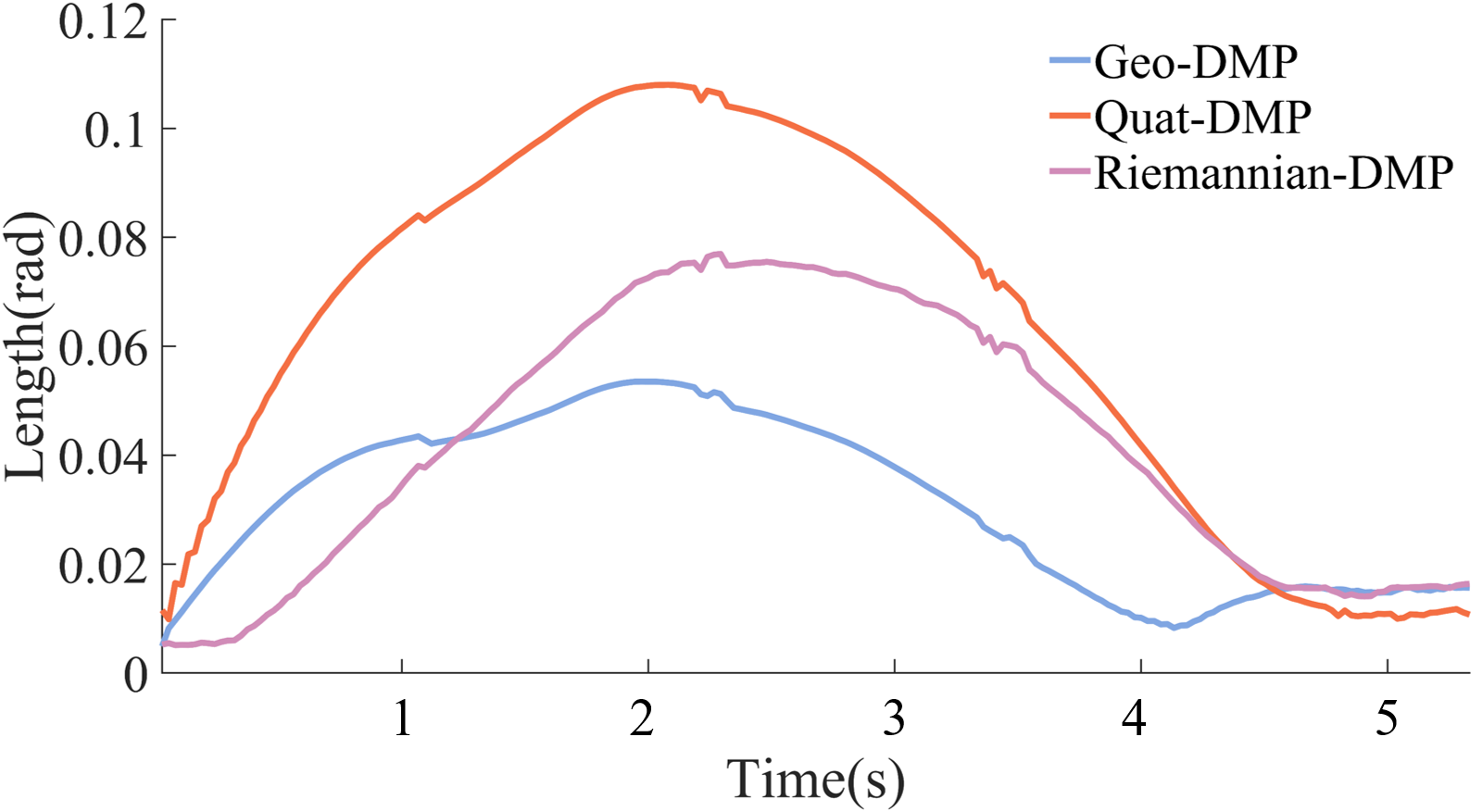}
    \caption{Comparison of errors between the proposed Geo-DMPs, Quat-DMPs, and Riemannian-DMPs.}
    \label{fig9}
\end{figure}

To analyze the errors of the Geo-DMPs, Quat-DMPs, and Riemannian-DMPs with the demonstration trajectory, the quaternion provided by (21) is used to calculate the distance of the geodesic on the ${{\mathcal{S}}^{3}}$ manifold. As shown in Fig. 9, compared to Quat-DMPs and Riemannian-DMPs, Geo-DMPs reduced the maximum error during the entire motion process by 61\% and 32\%, respectively. This is because Geo-DMP completely excludes the influence of time factors and relies solely on the nonlinear differential equation system of the motion space process to describe the motion. For machining tasks, the error during the motion process directly affects the outcome of the task. In terms of the final position of the orientation convergence, the convergence performance is worse than the Quat-DMPs because the Geo-DMPs exclude the time factor, and the velocity is not necessarily 0 after reaching the final position. In addition, since Geo-DMPs perform two derivations of the arc length, they are prone to oscillations if the sampling frequency is high; therefore, the sampling frequency of the demonstration needs to be reduced accordingly.

\subsubsection{Chamfer grinding Experiment}

To validate the synchronization accuracy of DMPs and their effectiveness in machining scenarios, a chamfer grinding experiment was designed. Two of the same C-shaped workpieces were customized and made from 6061 aluminum alloy. This material is known for its excellent mechanical properties and is commonly used in aerospace components such as aircraft frames and spacecraft bulkheads. Chamfer grinding skills were generated based on both Geo-DMPs and Quat-DMPs \cite{saveriano2023}, and positional trajectories were generated using AL-DMPs \cite{gaspar2018}.
\begin{figure}[h]
    \centering
    \includegraphics[scale=0.23]{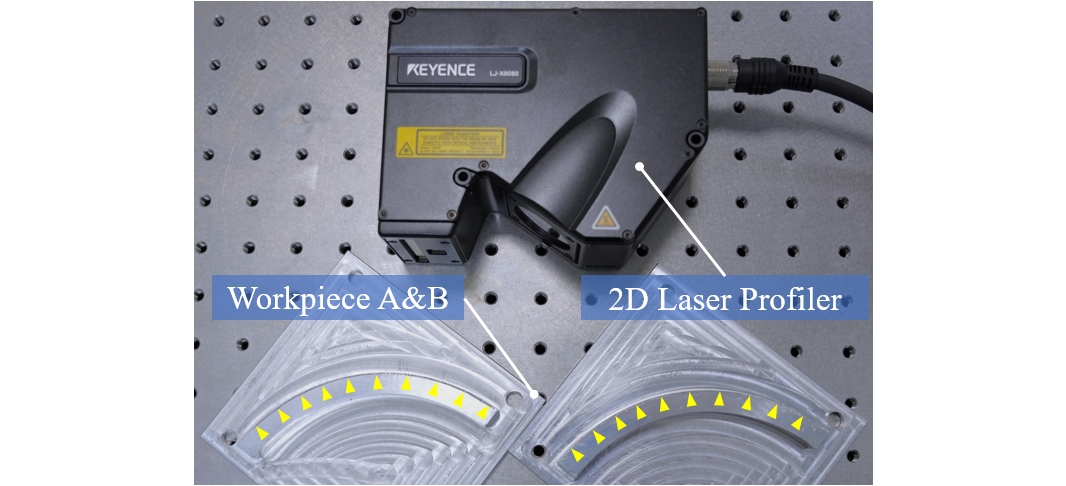}
    \caption{Setup of the line laser scanner for chamfer quality measurement.}
    \label{fig10}
\end{figure}
\begin{figure}[t]
    \centering
    \includegraphics[scale=0.21]{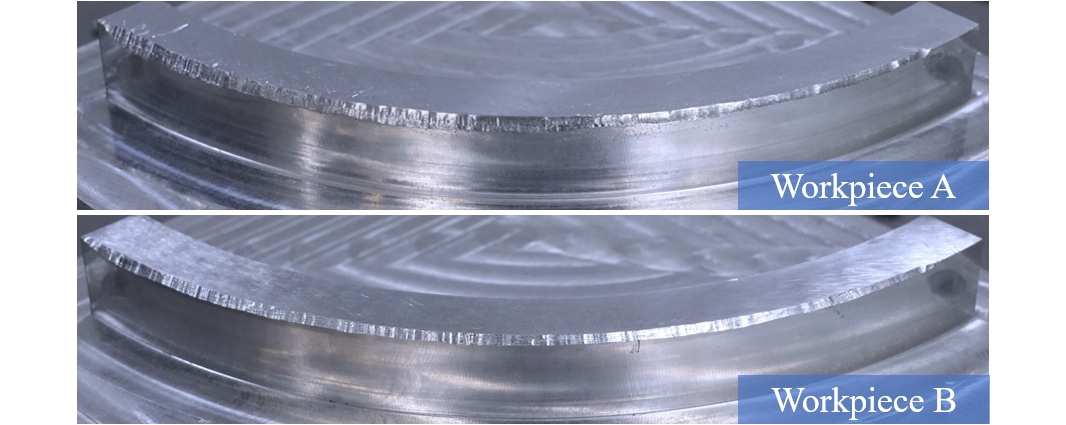}
    \caption{Evaluation of chamfer quality in experiments using DMPs synchronous encoding methods: Workpiece A was ground using Quat-DMPs, while Workpiece B was ground using Geo-DMPs.}
    \label{fig11}
\end{figure}
\begin{figure}[t]
    \centering
    \includegraphics[scale=0.14]{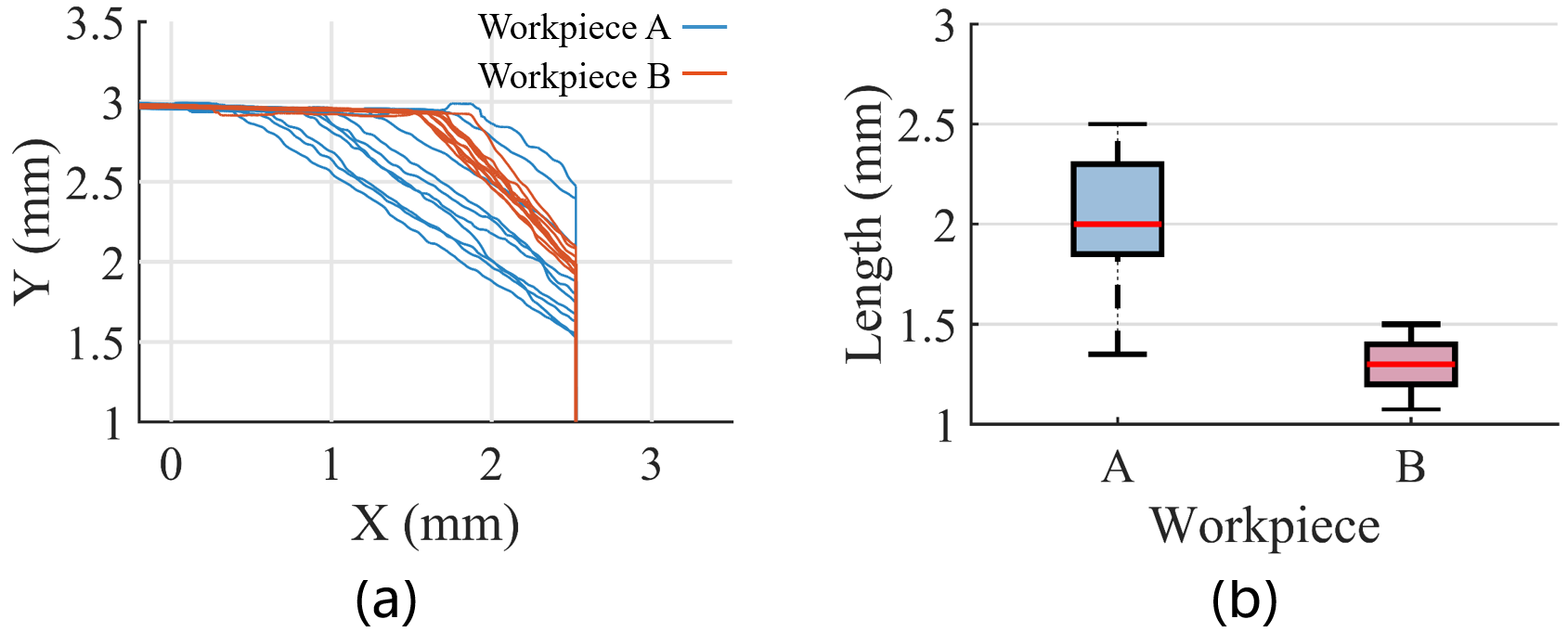}
    \caption{Scanned and statistical plots of workpiece chamfer contours. (a) The blue and red lines are the contours corresponding to the ten measurement points of workpiece A and workpiece B respectively. (b) The blue box plot on the left and the red box plot on the right represent the statistical results of the chamfer lengths of parts A and B, respectively.}
    \label{fig12}
\end{figure}

The grinding force was generated using the method described in Section III.C. The robot operated in impedance control mode with joint stiffness set to 1500 N/m. The quality of chamfer grinding is shown in Fig. \ref{fig11}, where Workpiece A was ground using Quat-DMPs, while Workpiece B was ground using Geo-DMPs. The Keyence LJ-X8060 line laser scanner was employed to scan the outer contour of the chamfer. The measurement setup is shown in Fig. \ref{fig10}, with the scanner positioned orthogonally to the chamfer. The measurement results are shown in Fig. \ref{fig12}.

Fig. \ref{fig12}(a) shows the workpiece profile after the chamfer grinding the chamfer angle generated by the kinematic skill based on the proposed method is close to 45 degrees and stable within this range, and the grinding error range is much smaller than that of the Quat-DMPs. The chamfer length statistics are shown in Fig. \ref{fig12}(b). The chamfer width was ground to a relatively precise range. The experimental result shows that the proposed method can generate high-precision robotic grinding trajectory.

\subsection{Free-form Surface Grinding}
 \begin{figure}[b]
    \centering
    \includegraphics[scale=0.24]{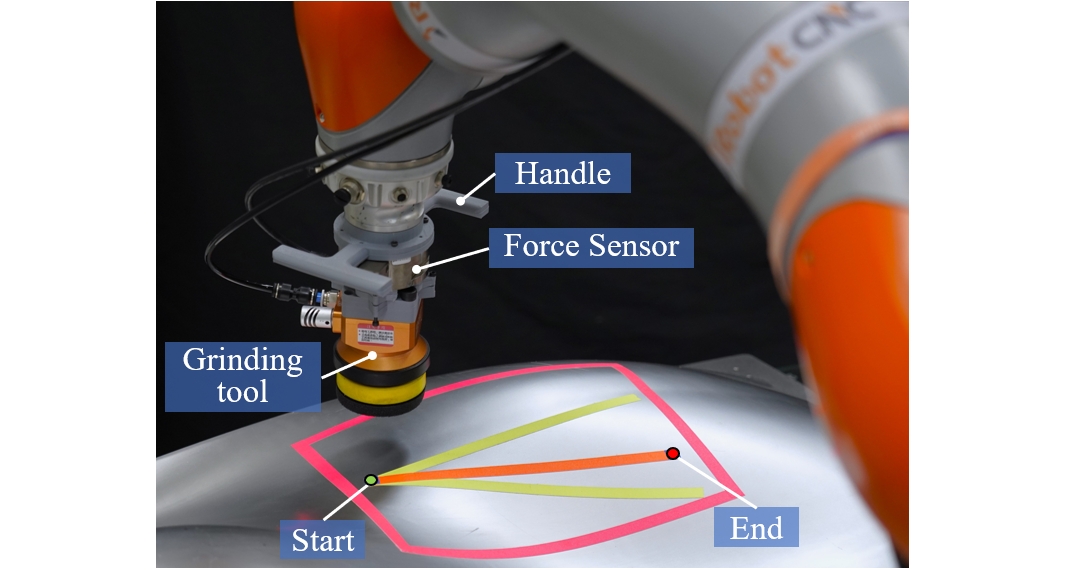}
    \caption{The robotic surface grinding platform showing the desired start and end points of the grinding trajectory on the surface.}
    \label{fig13}
\end{figure}
\subsubsection{Experiment of Surface Trajectory Generation}

In order to validate the effectiveness of the proposed method for generating grinding skills on surfaces, an experiment for generating surface trajectories was designed. The pneumatic grinding pen in Section IV.A was replaced with a pneumatic grinding disc, as shown in Fig. \ref{fig13}. A human operator demonstrated along the yellow trajectory, following the surface shape of the free-form surface. The robot recorded the end-effector position and orientation, which was then encoded using the method described in Section II. Subsequently, the operator specified the start and end points of a new trajectory, and the robot autonomously generated the grinding pose trajectory between them. The time series plot of the newly generated grinding skills is shown in Fig. \ref{fig15}. Using the encoding method described in Section II, the robot executed the grinding trajectory smoothly and uniformly.

\begin{figure*}[!t]
    \centering
    \includegraphics[scale=0.26]{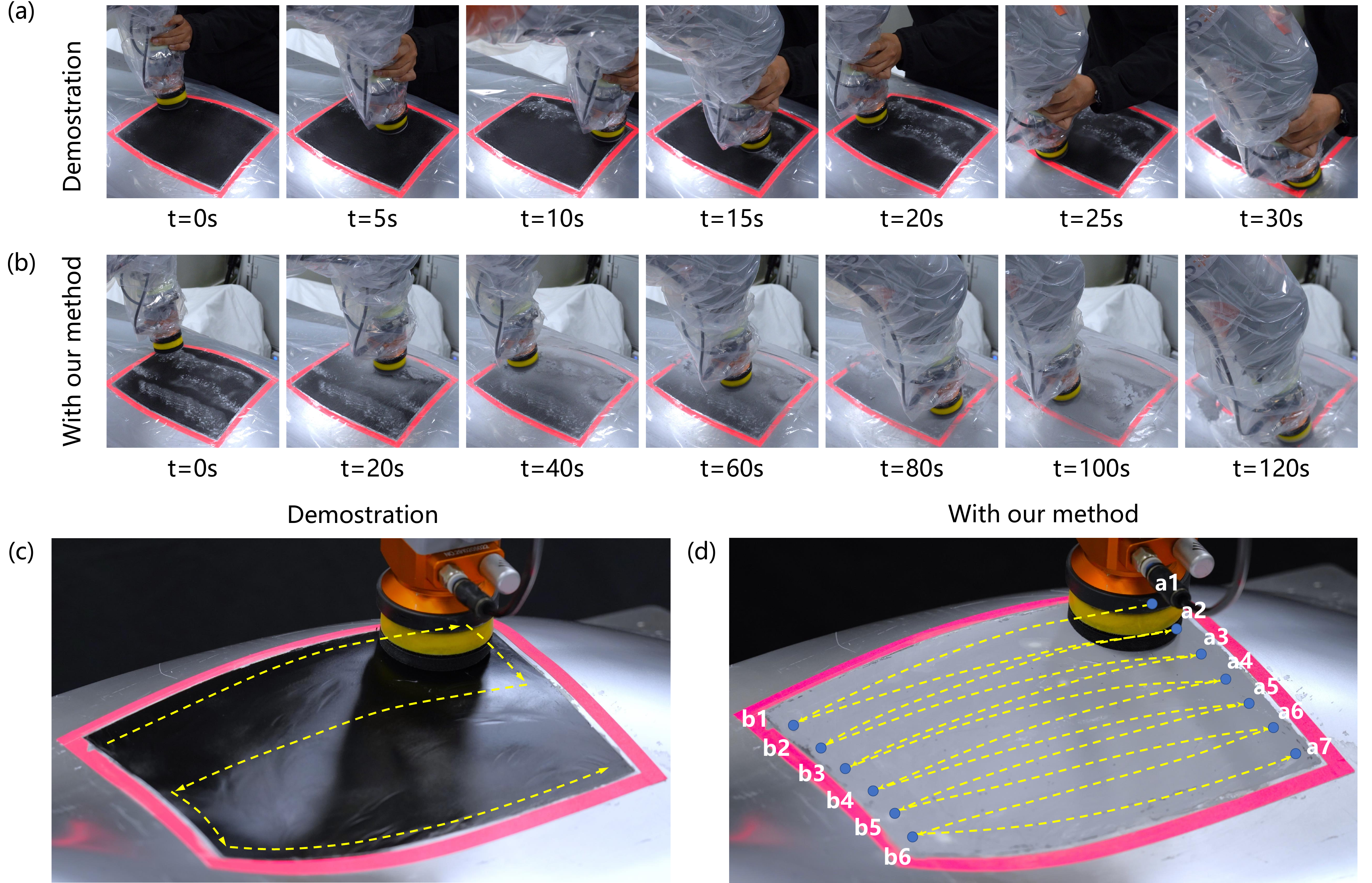}
    \caption{(a) Image sequence depicting the demonstrated robotic surface grinding trajectory. (b) Image sequence illustrating robotic surface grinding trajectory generated using DMPs based on a simple taught trajectory. (c) Schematic representation of the simple taught trajectory on the surface. (d) The grinding trajectories, which cover the entire surface, were generated using DMPs based on the taught trajectories.}
    \label{fig14}
\end{figure*}
\begin{figure}[h]
    \centering
    \includegraphics[scale=0.37]{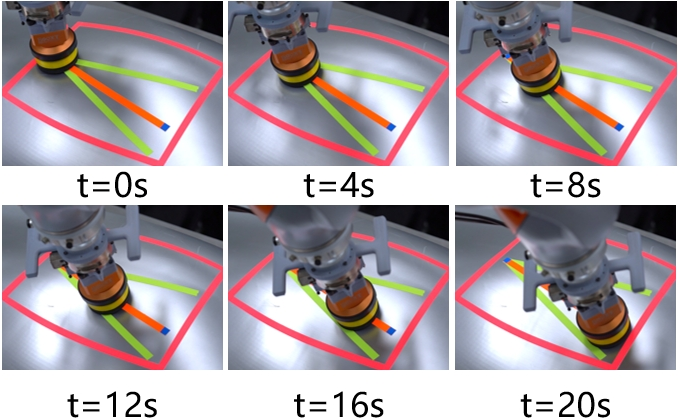}
    \caption{Image sequence depicting robotic grinding along an orange trajectory generated by DMPs, based on two fluorescent green trajectories demonstrated on the surface with specified start and end points.}
    \label{fig15}
\end{figure}

Due to the inability of existing methods to generate grinding trajectories with geometric constraints between any two points on free-form surfaces based on demonstration, the accuracy level was compared by directly contrasting the trajectories generated by this method with those generated from the design model, as shown in Fig. \ref{fig17}. It can be seen that the position error was under 3mm and the orientation error was under $8^{\circ}$.

\subsubsection{Surface Grinding Experiment}
We designed a free-form surface grinding experiment to test the effectiveness of the proposed method in real-world machining conditions. A layer of approximately 2.5mm thick grey putty, commonly used in automotive sheet metal, was applied to the curved surface. Then, a layer of black paint was sprayed onto the surface, with the left half having a thickness of about 2mm and the right half about 1mm. This setup was used to verify the method's ability to grind uneven coatings on curved surfaces.

As shown in Fig. \ref{fig14}(a), the first grinding demonstration was performed by a human craftsman. The robot was guided to roughly cover the area to be ground, outlined by the red boundary, and appropriate downward pressure was applied during the grinding process to ensure that the grinding disc could effectively remove the black paint layer. The pre-grinding effect of the free-form surface with black paint is shown in Fig. \ref{fig14}(c), with the reference trajectory of the teaching path marked by the yellow dashed lines. The free-form surface was then encoded using the method described in Section III, as shown by the contour of the free-form surface in Fig. \ref{fig17}(a). The encoding results of the grinding force distribution are shown in Fig. \ref{fig17}(b).
\begin{figure}[t]
    \centering
    \includegraphics[scale=0.20]{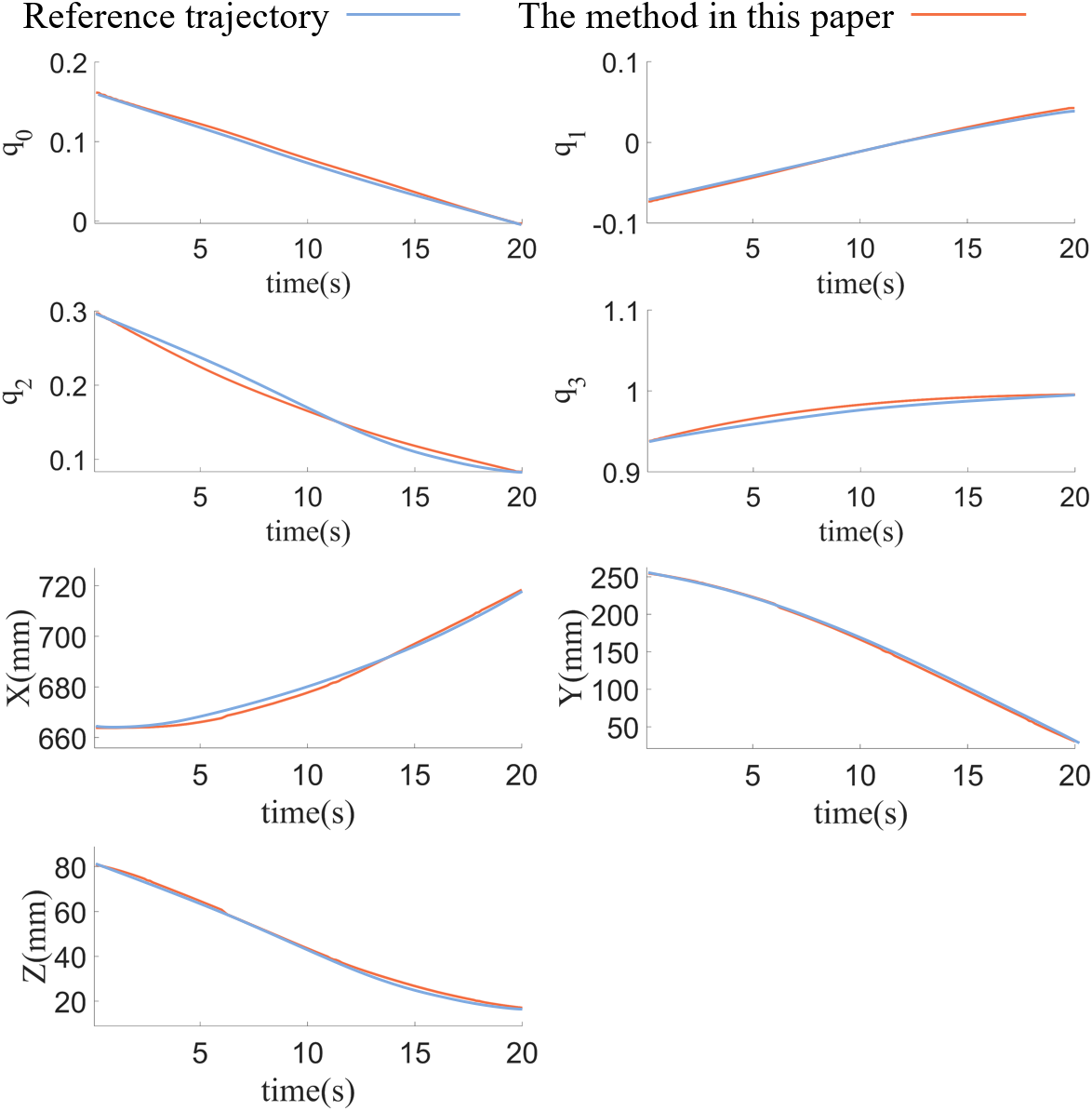}
    \caption{The trajectory generated by the proposed method is compared with the reference trajectory.}
    \label{fig16}
\end{figure}
\begin{figure}[h]
    \centering
    \includegraphics[scale=0.23]{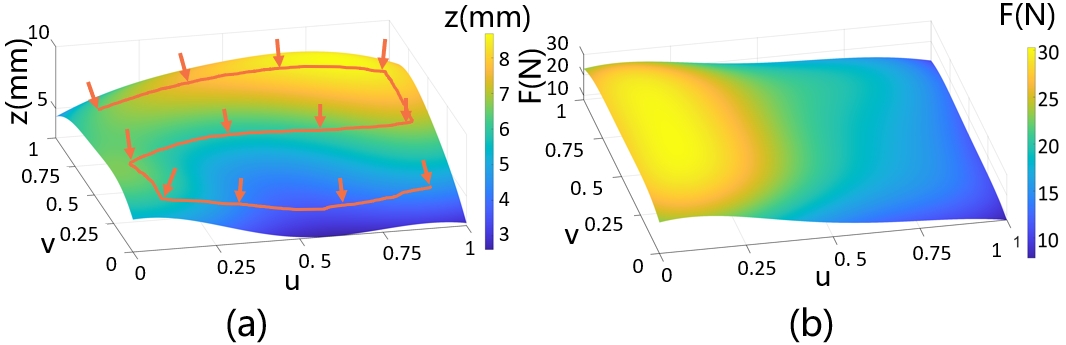}
    \caption{The method proposed in this paper has the following results on the encoding of surface grinding skills: (a) Encoding results for surface shapes. (b) Encoding results for grinding forces.}
    \label{fig17}
\end{figure}

A set of start and end points were defined on the curved surface: a1-a7, b1-b6. Each pair of points corresponds to the start and end points for generating fine-grinding reference trajectories, as indicated by the yellow dashed lines with arrows in Fig. \ref{fig14}(d). During this process, the robot applies greater grinding force in areas with thicker coatings to effectively remove the paint layer while using less grinding force in areas with thinner coatings to avoid wearing through the putty layer and damaging the workpiece. The execution time series of the skill is shown in Fig. \ref{fig14}(b), and the final grinding result is shown in Fig. \ref{fig14}(d) it can be seen that almost all of the black paint has been removed while the underlying putty layer remains intact. Five samples were cut from the ground surface and measured for roughness using a white light interferometer. The average roughness after grinding with 400-grit sandpaper was $S_a = 1.5 \, \mu m$.

\section{Conclusions}
In this article, we present a method for robotic grinding skills learning based on Geo-DMPs, enabling high-precision grinding skill generation on model-free surfaces. First, surface features are extracted from multiple grinding demonstration trajectories using intrinsic clustering algorithms and two-dimensional weighted Gaussian kernel function. Second, Geo-DMPs are proposed to achieve high-precision encoding of the robot's orientation with time-independent characteristics. Then, a framework for synchronously encoding robotic grinding position, orientation, and force is constructed, and synchronization is achieved using phase functions based on geodesic length. This framework not only enables high-precision generation of single grinding skill trajectories but also generates grinding skills between any two points on model-free surfaces.

A series of experiments were conducted to validate the effectiveness of the proposed method. The results demonstrated a significant improvement in accuracy compared to classical DMPs in orientation trajectory learning. In the experiment of model-free surface grinding, our method can autonomously generate complete grinding skills for the surface with only the demonstration provided by human craftsmen. This method holds substantial applicability in real-world industrial settings.

In future work, we aim to incorporate feedback on processing effects through vision and explore how to integrate human closed-loop skill generation strategies into the construction of robot measurement-processing closed loops. This will further enhance the capabilities of robots in industrial scenarios.


\vspace{-0.3in}
\begin{IEEEbiography}[{\includegraphics[width=1.05in,height=1.25in,clip,keepaspectratio]{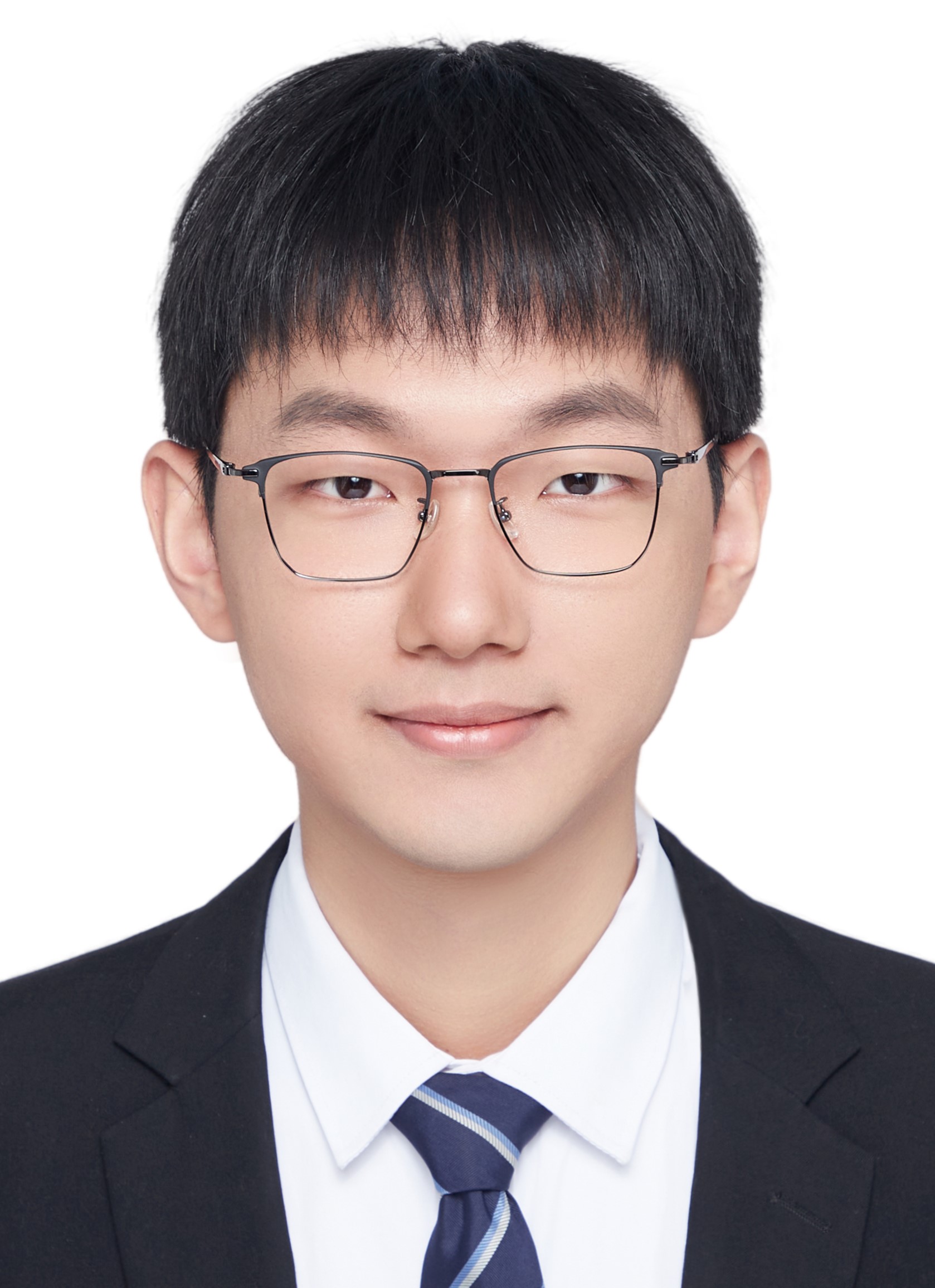}}]{Shuai Ke}
received the B.E. degree from the School of Automation, China University of Geosciences, Wuhan, China, in 2023. He is pursuing a PhD in mechanical engineering with the Huazhong University of Science and Technology, Wuhan, China. His research interests include perception, learning and control of robots and robotic machining.
\end{IEEEbiography}
\vspace{-0.3in}
\begin{IEEEbiography}[{\includegraphics[width=0.94in,height=1.25in,clip,keepaspectratio]{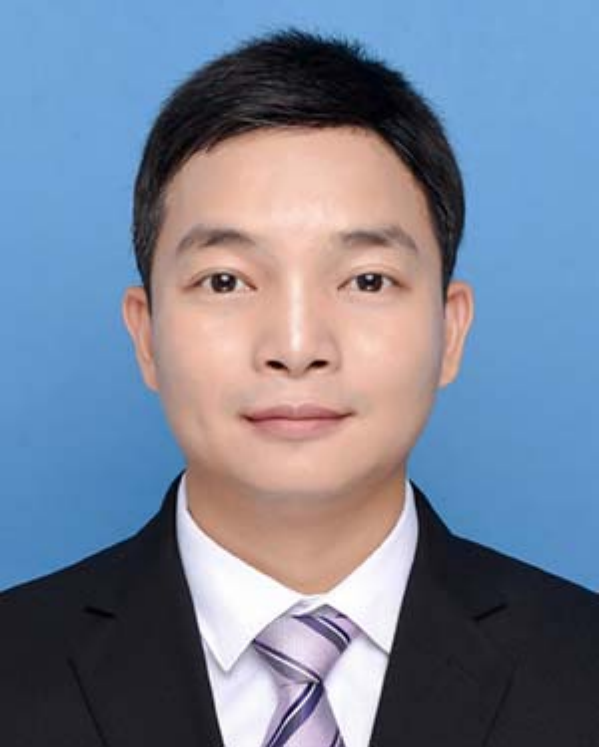}}]{Huan Zhao}
(Member, IEEE) received the B.S.
degree in mechanical engineering from Jilin University, Changchun, China, in 2006, and
the Ph.D. degree in mechanical engineering from Shanghai Jiao Tong University, Shanghai,
China, in 2013.
He conducted postdoctoral research with the Huazhong University of Science and Technology, Wuhan, China, in 2013, where he is currently a Professor. His research interests include force control, visual servoing, and machine learning with applications to robotic machining
\end{IEEEbiography}
\vspace{-0.4in}
\begin{IEEEbiography}[{\includegraphics[width=1in,height=1.25in,clip,keepaspectratio]{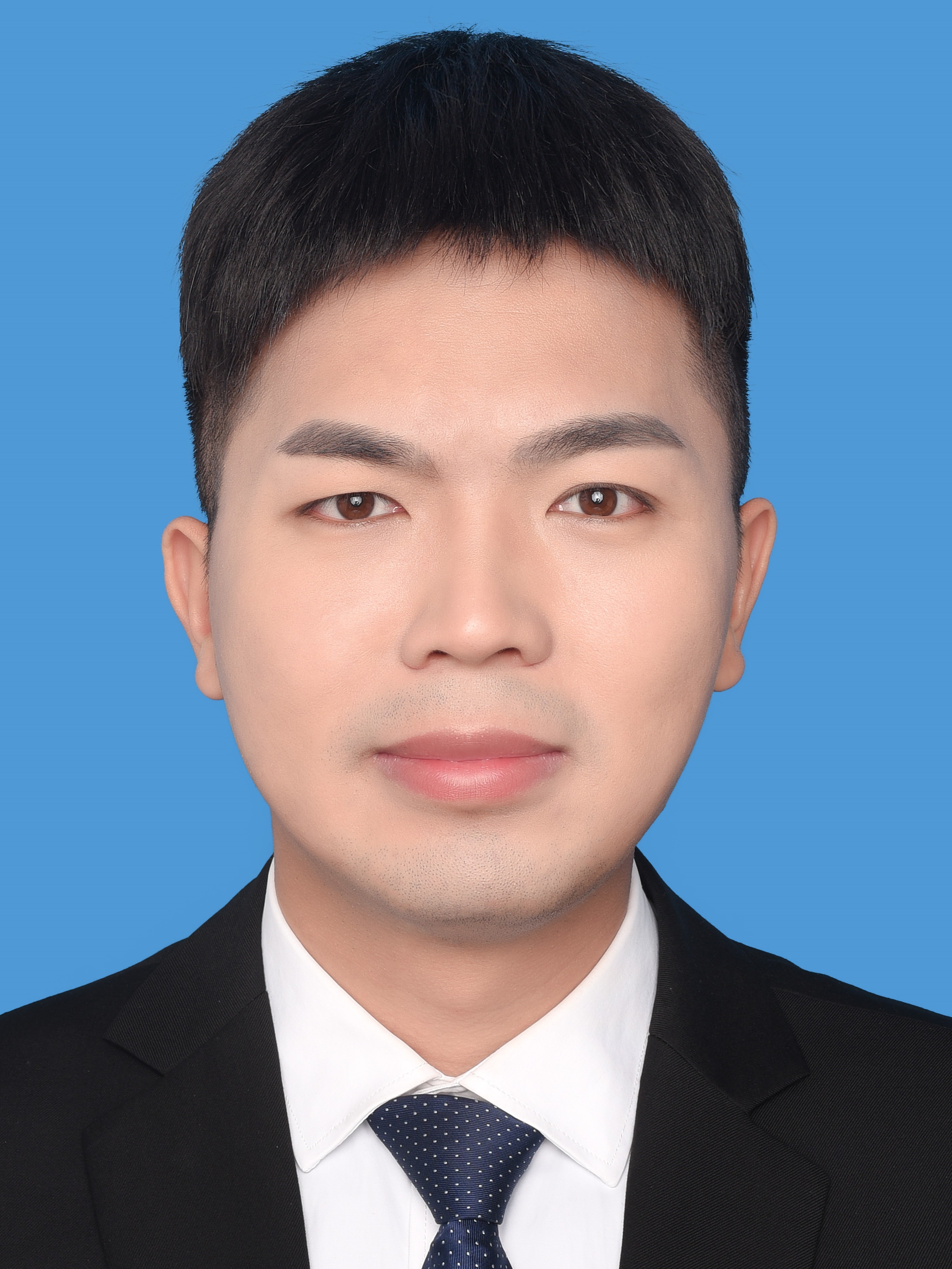}}]{Xiangfei Li}
received the B.E. degree from the School of Mechanical Science and Engineering, Jilin University, Changchun, China, in 2012, and the Ph.D. degree from the School of Mechanical Science and Engineering, Huazhong University of Science and Technology, Wuhan, China, in 2020. He is currently an Assistant Research Fellow with the Huazhong University of Science and Technology. His research interests include visual servoing, skills transferring and collaborative compliance for robotic assembly.
\end{IEEEbiography}
\vspace{-0.38in}
\begin{IEEEbiography}[{\includegraphics[width=1in,height=1.25in,clip,keepaspectratio]{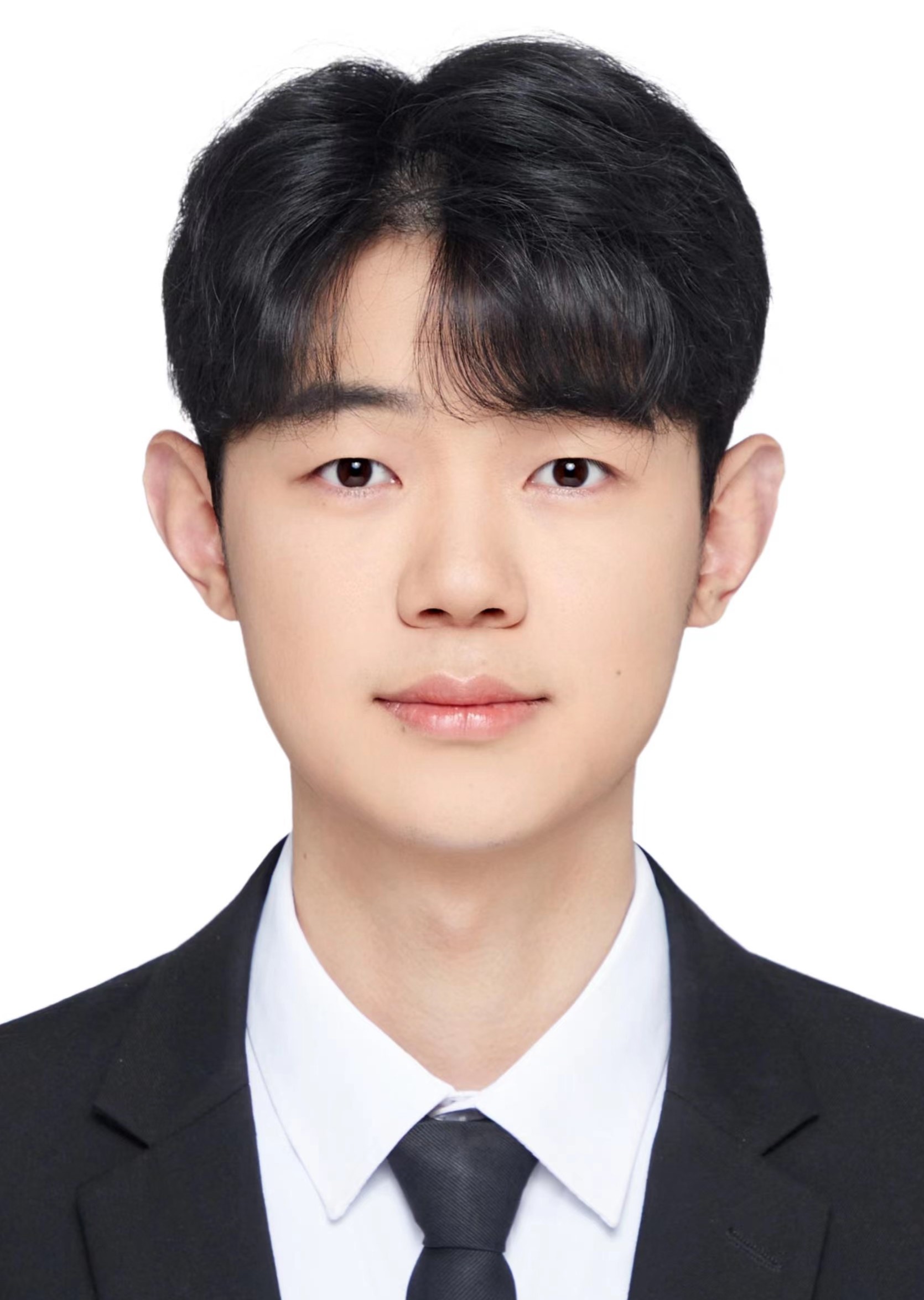}}]{Zhiao Wei}
received the B.E. degree in mechanical engineering from Nanjing University of Aeronautics and Astronautics, Nanjing, China, in 2023. He is currently pursuing an M.S. degree in mechanical engineering at Huazhong University of Science and Technology, Wuhan, China. His research interests include perception, learning and control of robots and robotic machining.
\end{IEEEbiography}
\vspace{-0.3in}
\begin{IEEEbiography}[{\includegraphics[width=1.1in,height=1.25in,clip,keepaspectratio]{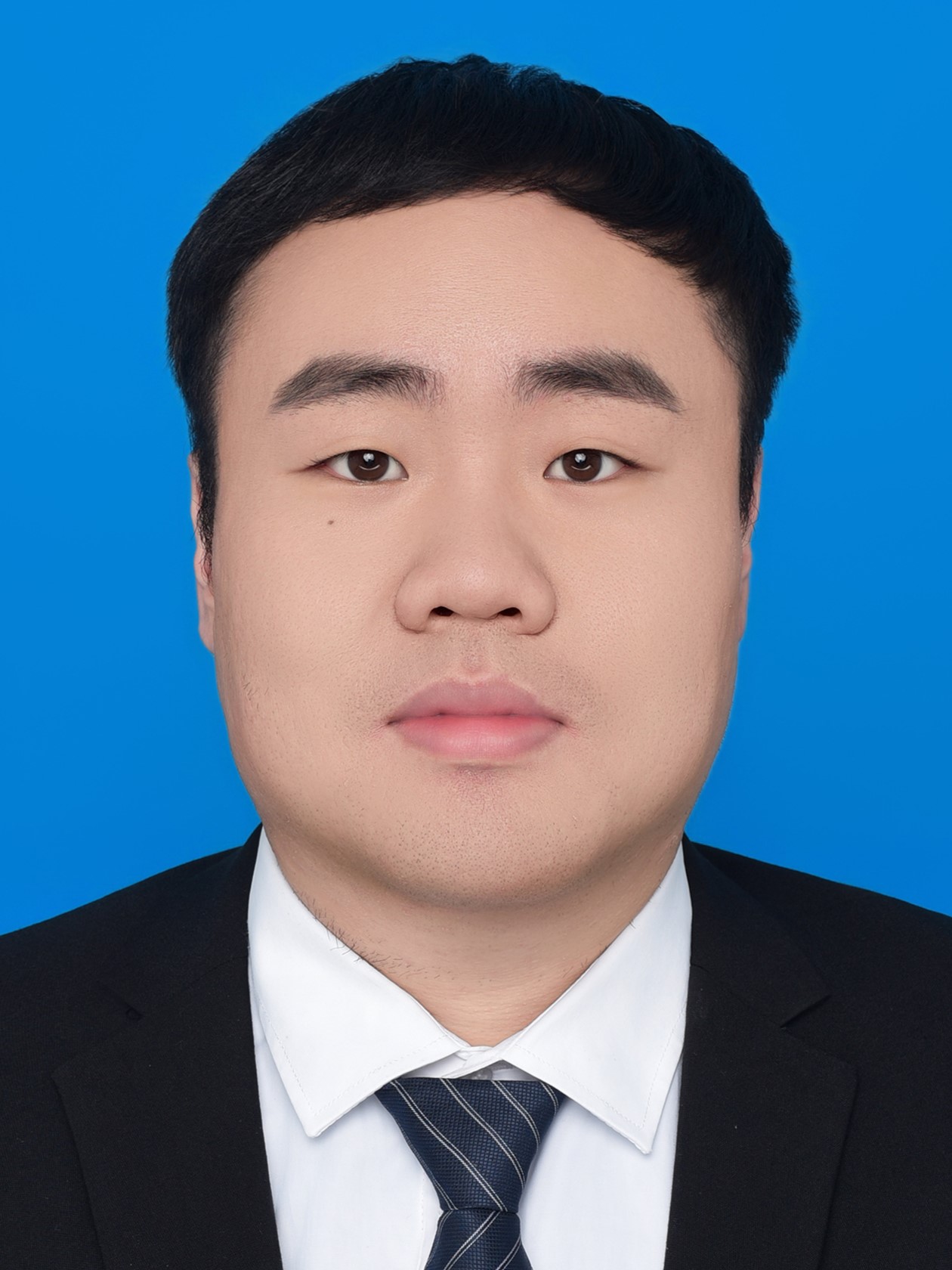}}]{Yecan Yin}
received his B.E. degree from Huazhong University of Science and Technology in 2020, Wuhan, Hubei, China. He is now pursuing his PhD degree in Huazhong University of Science and Technology. His research interests include visual servoing, nonlinear system, computer vision.
\end{IEEEbiography}
\vspace{-0.3in}
\begin{IEEEbiography}[{\includegraphics[width=1in,height=1.25in,clip,keepaspectratio]{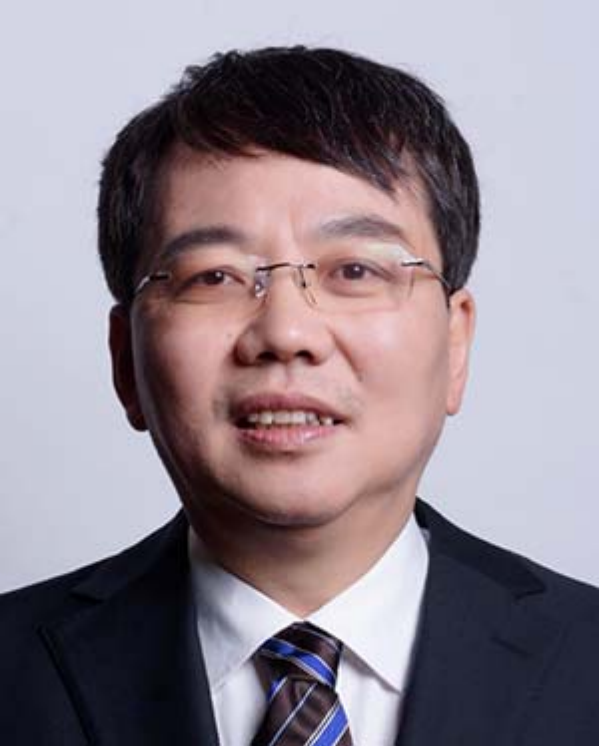}}]{Han Ding}
(Senior Member, IEEE) received the Ph.D. degree in mechanical engineering from the Huazhong University of Science and Technology (HUST), Wuhan, China, in 1989.
He has been a Professor with the HUST, since 1997. He was a Cheung Kong Chair Professor with Shanghai Jiao Tong University, Shanghai, China, from 2001 to 2006. His research interests include robotics, multiaxis machining, and control engineering.
Dr. Ding was elected a Member of the Chinese Academy of Sciences, in 2013.
\end{IEEEbiography}

\vfill

\end{document}